\documentclass{report}
\usepackage{setspace}
\usepackage[unicode]{hyperref}
\usepackage{amssymb,graphicx,color}
\usepackage{booktabs}
\usepackage{amsfonts}
\usepackage{latexsym}
\usepackage{amsmath}
\usepackage{tabularx}
\usepackage{multirow}
\usepackage{makecell} 
\usepackage{algorithm}
\usepackage{algorithmic}
\usepackage{subcaption}
\usepackage{siunitx}
\usepackage[table]{xcolor}
\definecolor{lightgray}{gray}{0.9}  
\definecolor{highlight}{RGB}{255, 200, 200}  
\newcolumntype{C}{>{\centering\arraybackslash}X}

\newcommand{\SA}[1]{\leavevmode \textcolor{magenta}{#1}}
\renewcommand{\SA}[1]{}

\usepackage{microtype}

\title{  	
{{\Huge Neural PDE Solvers with Physics Constraints}}\\
{\large A Comparative Study of PINNs, DRM, and WANs}\\
		}
\date{}
\author{Jiakang Chen\thanks{
{\bf Disclaimer:}
This report is submitted as part requirement for the MSc Machine Learning at UCL. It is
substantially the result of my own work except where explicitly indicated in the text. The report may be freely copied and distributed provided the source is explicitly acknowledged
}
\\ \\
Supervisor: Simon Arridge}

\begin{document}
 \onehalfspacing
\maketitle

\begin{abstract}
Partial differential equations (PDEs) underpin models across science and engineering, yet analytical solutions are atypical and classical mesh-based solvers can be costly in high dimensions. This dissertation presents a unified comparison of three mesh-free neural PDE solvers, physics-informed neural networks (PINNs), the deep Ritz method (DRM), and weak adversarial networks (WANs), on Poisson problems (up to 5D) and the time-independent Schr\"odinger equation in 1D/2D (infinite well and harmonic oscillator), and extends the study to a laser-driven case of Schrödinger's equation via the Kramers-Henneberger (KH) transformation.

Under a common protocol, all methods achieve low $L_2$ errors ($10^{-6}$-$10^{-9}$) when paired with forced boundary conditions (FBCs), forced nodes (FNs), and orthogonality regularization (OG). Across tasks, PINNs are the most reliable for accuracy and recovery of excited spectra; DRM offers the best accuracy-runtime trade-off on stationary problems; WAN is more sensitive but competitive when weak-form constraints and FN/OG are used effectively. Sensitivity analyses show that FBC removes boundary-loss tuning, network width matters more than depth for single-network solvers, and most gains occur within 5000-10,000 epochs. The same toolkit solves the KH case, indicating transfer beyond canonical benchmarks.

We provide practical guidelines for method selection and outline the following extensions: time-dependent formulations for DRM and WAN, adaptive residual-driven sampling, parallel multi-state training, and neural domain decomposition. These results support physics-guided neural solvers as credible, scalable tools for solving complex PDEs.
\end{abstract}

\tableofcontents

\clearpage\setcounter{page}{1}

\chapter{Introduction}
Partial differential equations (PDEs) are the lingua franca of continuum models across mathematics, physics, and engineering. Classical examples include heat conduction governed by the heat equation~\cite{cannon_one-dimensional_1984}, fluid motion described by the Navier-Stokes equations~\cite{mclean_understanding_2012}, electromagnetic waves captured by Maxwell’s equations~\cite{hampshire_derivation_2018}, and structural deformation modelled by the equations of elasticity. By encoding the dynamics of fields and forces, PDEs enable the prediction and optimization of system behaviour under varied initial and boundary conditions.

Exact analytical solutions are available only for special cases so numerical approximation is indispensable. The finite difference method (FDM) replaces derivatives with difference quotients on structured grids~\cite{ozisik_finite_2017}; the finite volume method (FVM) enforces conservation over control volumes~\cite{leveque_finite_2002}; spectral and spectral element methods attain high-order accuracy for smooth solutions via global polynomial expansions~\cite{komatitsch_introduction_1999}; and the finite element method (FEM), developed in the mid–twentieth century~\cite{rai_dynamics_2002}, uses unstructured meshes to handle complex geometries and heterogeneous media with great flexibility.

Despite their success, mesh-based techniques face persistent hurdles in modern applications. Generating high-quality meshes in high dimensions is costly: complex geometries often require substantial manual intervention, and degrees of freedom grow rapidly under refinement. These challenges have motivated mesh-free alternatives.

Recent advances in deep learning have produced neural PDE solvers that embed physical laws directly into training objectives. Physics-informed neural networks (PINNs) penalize the PDE residual and boundary conditions in the loss so that the learned surrogate satisfies the governing equations. The deep Ritz method (DRM) minimizes the variational energy functional associated with elliptic problems, providing a principled weak formulation. Weak adversarial networks (WANs) cast solution and test functions as a generator-discriminator pair that optimizes a weak form adversarially. By incorporating physics constraints, these approaches can reduce data demands and avoid meshing, while remaining flexible across problem families. At the same time, they introduce distinct optimization pathologies (for example, spectral bias or adversarial instability) and design choices (loss weighting, sampling, and architecture) that materially affect accuracy and computational cost.

This dissertation undertakes a systematic study of three neural solvers, PINN, DRM, and WAN, on canonical elliptic and quantum benchmarks. First, we compare their performance on the Poisson equation and on the time-independent Schr\"odinger equation (infinite potential well and harmonic oscillator) in one and two spatial dimensions under a unified evaluation protocol. Second, we analyze and improve training through principled modifications, including forced boundary conditions (FBC) that enforce boundaries by construction, orthogonality-promoting penalties for excited states, and the use of fixed or forced nodes where nodal sets are known. Third, we extend the study beyond stationary settings by formulating and solving the laser-driven Schr\"odinger equation in the Kramers-Henneberger (KH) frame with neural methods, using it as a stress test of robustness and physical fidelity. Throughout, we report sensitivity analyses, ablations, and reproducibility checks, and we distil practical guidance for method selection and configuration.

The dissertation is organized as follows. Chapter~2 reviews classical numerical methods and emerging neural network-based approaches for PDEs, situating the present work in the literature. Chapter~3 lays out the methodological framework for PINNs, DRM, and WANs, specifies the benchmark problems and evaluation metrics, and details architectures, sampling, and optimization. Chapter~4 presents empirical results for the Poisson and stationary Schr\"odinger benchmarks, including both baseline implementations and targeted improvements. Chapter~5 formulates the laser-driven Schr\"odinger equation in the KH frame and develops a neural solver for the time-dependent setting. Chapter~6 synthesizes the findings, compares methods across tasks, discusses limitations and threats to validity, and outlines future directions. Chapter~7 concludes.

\chapter{Literature Review}
Partial Differential Equations (PDEs) are widely used to formulate complex problems across the sciences. In this chapter, we will outline the basic PDE concepts before reviewing the classical numerical solution methods and their limitations. Then, we will move to the foundations of neural network and deep learning theory as, using the differentiation and optimization method, neural networks give a perfect platform for solving PDEs. Neural network-based PDE solvers will be discussed at the end of this chapter to ascertain the  history and state of the art of this field. We also investigate their application to solving Schrödinger problems.

\section{PDE Fundamentals}

Partial differential equations (PDEs) are equations in which the unknown function \(u\) depends on multiple independent variables (typically spatial coordinates \(\mathbf{x}\in\Omega\subset\mathbb{R}^d\) and possibly time \(t\)), and the equation involves partial derivatives of \(u\).  A general second-order PDE can be written in the form
\begin{equation}
    F\bigl(\mathbf{x},t,\,u,\,\nabla u,\,\nabla^2 u\bigr) \;=\; 0,
\end{equation}
where \(\nabla u\) and \(\nabla^2 u\) denote the gradient and Hessian (matrix of second derivatives) of \(u(\mathbf{x},t)\). More generally, we can write PDEs of the $k$th order as
\begin{equation}
    F\bigl(\mathbf{x},t,\,u,\,D u,\,D^2 u, \cdots , D^ku\bigr) \;=\; 0,
\end{equation}
where $D^n$ is partial derivative operator.
\subsection{Classification of PDEs}
PDEs are called \emph{linear} if they are linear in $u$ and its derivatives. Second-order linear PDEs, with unknown function $u(x,y)$, can be written in the general form
\begin{equation}
a_1 u_{x x}+a_2 u_{x y}+a_3 u_{y x}+a_4 u_{y y}+a_5 u_x+a_6 u_y+a_7 u=f,
\end{equation}
where $a_i$ and f are functions of $x$ and $y$ only. 

A linear PDE is deemed \emph{homogeneous} when $f$ is zero everywhere. 

We define a \emph{quasilinear} PDE as when $a_i$ and $f$ are functions of unknown and lower-order derivatives, and the highest order derivatives appear only as linear. For example, Einstein's equations of general relativity form a quasilinear PDE. 

Finally, we have \emph{fully nonlinear} PDEs which have nonlinearities on one or more of the highest-order derivatives. One well-known example is the Monge-Ampere equation~\cite{mooney_monge-ampere_2018} in differential geometry. \SA{Provide a reference and/or write down the equation(s)}

In our discussion, we will focus on second-order linear PDEs. These can be further classified to help us to understand their initial and boundary conditions. They can be classified by the sign of the eigenvalues of the coefficient matrix \(A(\mathbf{x})\) in the principal part
\begin{equation}
    \sum_{i,j=1}^d A_{ij}(\mathbf{x})\,\frac{\partial^2 u}{\partial x_i\,\partial x_j}.
\end{equation}

\begin{itemize}
  \item \textbf{Elliptic:} \(\lambda_i\) all of same sign. E.g., Laplace’s equation\footnote{Here, \(\Delta u\) denotes the Laplacian of \(u\), which is the trace of the Hessian matrix.} \(\Delta u = 0\)
, Poisson’s equation \(\Delta u = f\).  Solutions are generally smooth and boundary‐value problems are well‐posed.
  \item \textbf{Parabolic:} one zero eigenvalue, rest of the same sign. E.g., the heat equation \(u_t - \Delta u = 0\).  These describe diffusion‐like processes and require both initial and boundary data.
  \item \textbf{Hyperbolic:} mixed signs. E.g., the wave equation \(u_{tt} - c^2\Delta u = 0\).  These model propagation phenomena and require Cauchy data (initial displacement and velocity) plus boundary conditions if the domain is bounded.
\end{itemize}

\subsection{Boundary and Initial Value Problems}

For a problem to be well posed, good initial conditions (ICs) and boundary conditions (BCs) are important. If we have too many conditions, we will not find a solution. If the conditions are weak and too few, we will get non-unique solutions. Sometimes, small errors in ICs or BCs will cause large changes in the solution. Here are some common choices on ICs and BCs.
\paragraph{Boundary Conditions.}

On a spatial boundary \(\partial\Omega\), one commonly prescribes the following boundary conditions:
\begin{itemize}
  \item Dirichlet: \(u(\mathbf{x},t) = g_D(\mathbf{x},t)\) for \(\mathbf{x}\in\partial\Omega\).
  \item Neumann: \(\nabla u(\mathbf{x},t)\cdot\mathbf{n} = g_N(\mathbf{x},t)\), where \(\mathbf{n}\) is the outward normal.
  \item Robin (mixed): \(\alpha u + \beta\,\nabla u\cdot\mathbf{n} = g_R\).
\end{itemize}

\paragraph{Initial Conditions.}
For time–dependent (parabolic or hyperbolic) PDEs, one must also specify
\[
u(\mathbf{x},0) = u_0(\mathbf{x}).
\quad
\]

In order to compare the neural network-based method, we will mainly focus on the time-independent case with Dirichlet boundary conditions. 

\section{Classical Numerical Methods}
Not all the PDEs have exact solutions, in which case computational approximations are sought. Classical numerical methods for solving PDEs approximate the continuous problem by a discrete system that can be solved on a computer.  The three most common approaches are finite difference methods (FDMs), finite element methods (FEMs), and spectral methods.

\subsection{Finite Difference Methods}
The finite difference methods (FDMs) are a powerful class of numerical techniques for solving PDEs by replacing continuous derivatives with discrete difference quotients. In essence, the spatial and temporal domains of a PDE are overlaid with a grid of points, and the derivatives at each point are approximated by algebraic combinations of neighbouring values. This process transforms the original problem into a system of linear (or non-linear) equations that can be efficiently marched forward in time or solved iteratively until they converge on a steady state. 

The difference quotients can be derived from Taylor's theorem. First, we expand the function $f(x)$ as: 
\begin{equation}
f\left(x_0+h\right)=f\left(x_0\right)+\frac{f^{\prime}\left(x_0\right)}{1!} h+\frac{f^{(2)}\left(x_0\right)}{2!} h^2+\cdots+\frac{f^{(n)}\left(x_0\right)}{n!} h^n+R_n(x),
\end{equation}
where $R_n(x)$ is a remainder term. Now, we can approximate the first derivative as:
\begin{equation}
f^{\prime}\left(x_0\right)=\frac{f\left(x_0+h\right)-f\left(x_0\right)}{h}-\frac{R_1(x)}{h} .
\end{equation}

\begin{figure}[ht]
  \centering
  \includegraphics[width=0.7\textwidth]{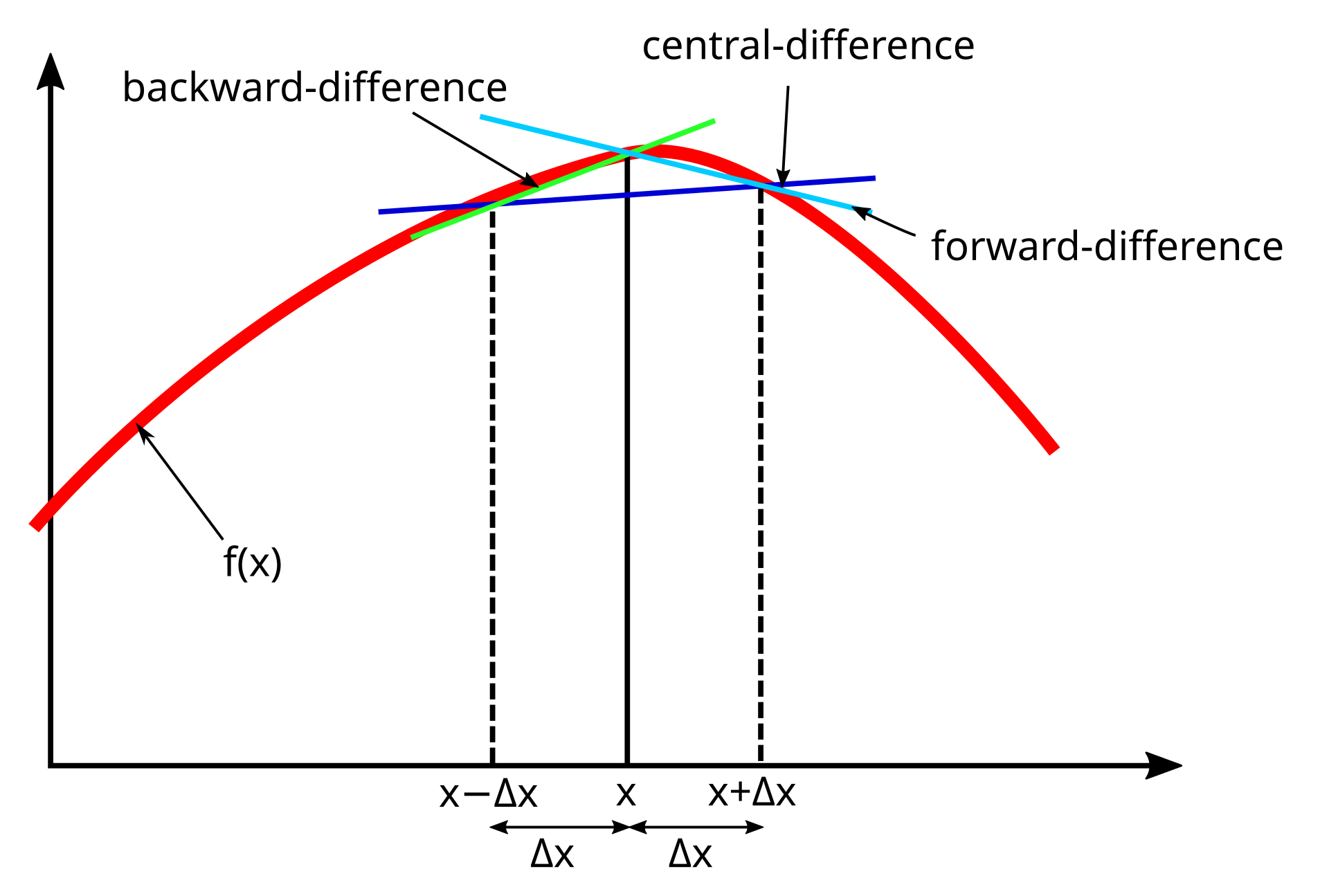}
  \caption{Finite‐difference approximations at $x=0$: forward, backward, and central differences~\cite{kakitc_finite_2025}.}
  \label{fig:fd_tangents}
\end{figure}

\begin{figure}[ht]
  \centering
  
  \begin{subfigure}[b]{0.3\textwidth}
    \centering
    \includegraphics[width=\textwidth]{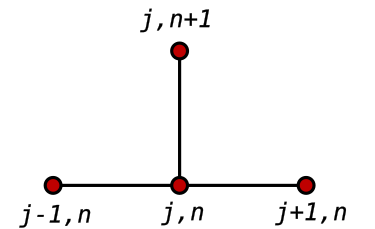}
    \caption{Explicit (forward Euler)}
    \label{fig:explicit}
  \end{subfigure}%
  \hfill
  \begin{subfigure}[b]{0.3\textwidth}
    \centering
    \includegraphics[width=\textwidth]{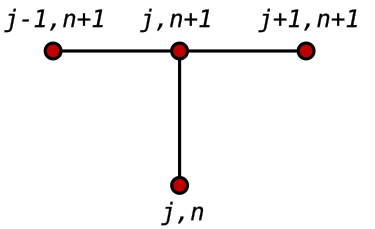}
    \caption{Implicit (backward Euler)}
    \label{fig:implicit}
  \end{subfigure}%
  \hfill
  \begin{subfigure}[b]{0.3\textwidth}
    \centering
    \includegraphics[width=\textwidth]{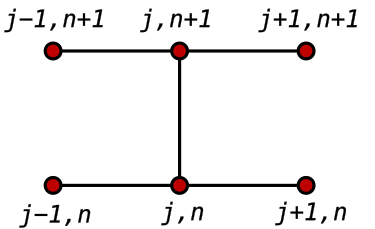}
    \caption{Crank-Nicolson}
    \label{fig:crank_nicolson}
  \end{subfigure}
  
  \caption{Comparison of time-stepping schemes: explicit, implicit, and Crank-Nicolson~\cite{kakitc_finite_2025}.}
  \label{fig:time_schemes}
\end{figure}

The truncation error, $R_1(x)/h$, typically scales with $h$ (first-order) but, by combining Taylor expansions at points $x_0\pm h$, one can derive central-difference schemes with $\mathcal{O}(h^2)$ accuracy. Similar definitions extend to higher derivatives and multiple dimensions. In practice, we need to consider consistency, stability and convergence when using FDMs. Thus, we have different choice of stencil (e.g., forward, backward or central differences) shown in Fig \ref{fig:fd_tangents} and time‐stepping scheme (e.g., explicit, implicit or Crank-Nicolson) shown in Fig \ref{fig:time_schemes}. 

Building on the Taylor‐expansion derivation and our choice of difference approximations, finite‐difference methods remain conceptually straightforward and trivial to implement on structured (Cartesian) grids. By widening the stencil, for example, moving from a three-point to a five-point, seven-point, or even compact scheme, the formal order of spatial accuracy can be raised (for instance, from second- to fourth- or sixth-order) without altering the mesh itself.

However, that very simplicity becomes a liability on complex or unstructured domains. To fit an irregular boundary into a Cartesian layout, one must introduce coordinate transformations or ghost-point interpolations, which, while practical, often erode the scheme’s formal convergence rate and can undermine stability when enforcing sophisticated boundary conditions.
\subsection{Finite Element Method}
\begin{figure}[ht]
  \centering
  \begin{subfigure}[b]{0.40\textwidth}
    \centering
    \includegraphics[width=\textwidth]{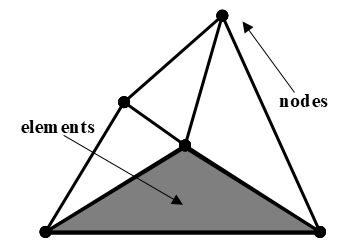}
    \caption{2D triangular mesh showing elements and nodes.}
    \label{fig:mesh}
  \end{subfigure}\hfill
  \begin{subfigure}[b]{0.45\textwidth}
    \centering
    \includegraphics[width=\textwidth]{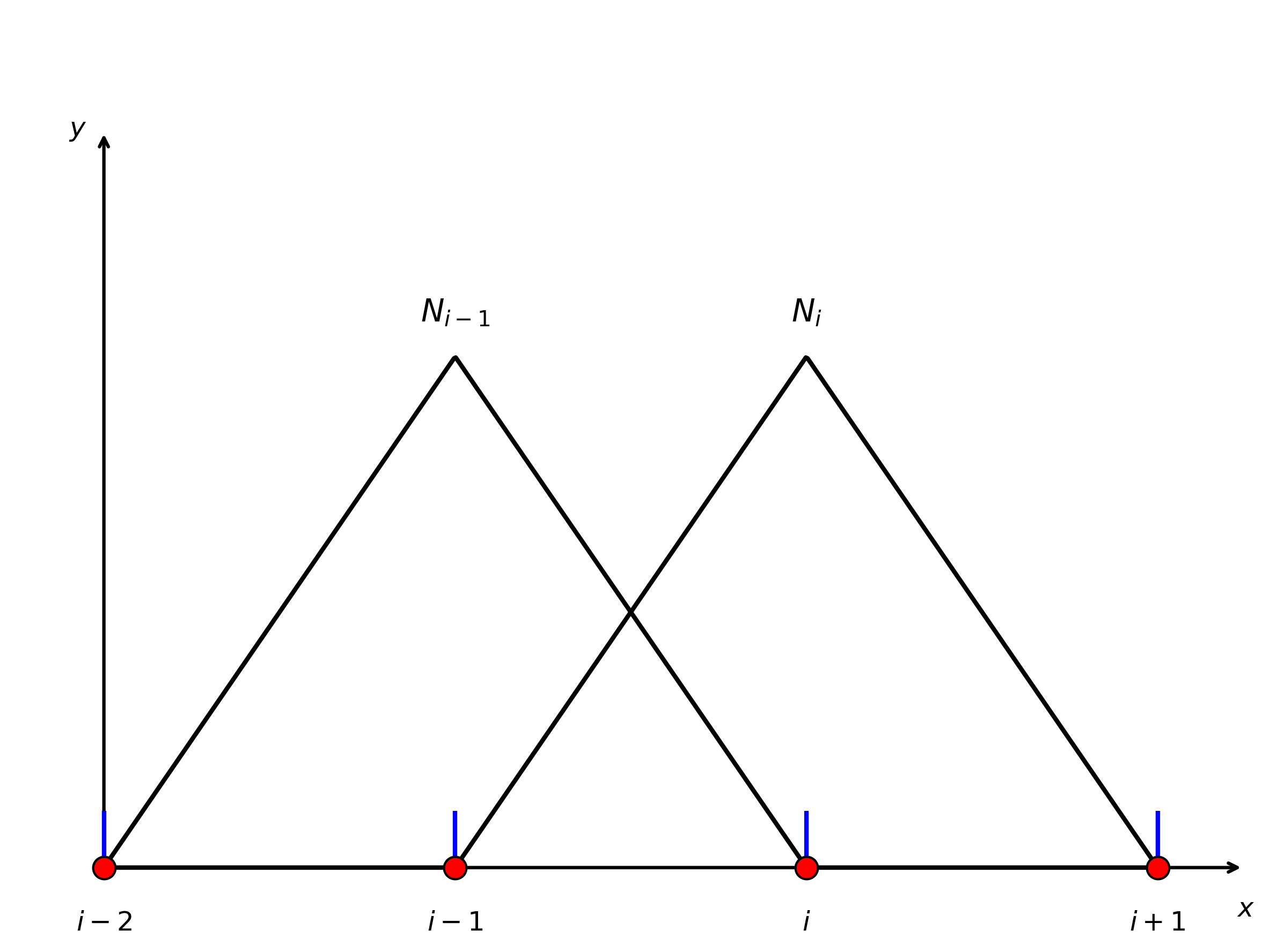}
    \caption{1D linear shape functions $N_{i-1}(x)$ and $N_i(x)$ over a uniform mesh.}
    \label{fig:shape}
  \end{subfigure}
  \caption{(a) Discretization of domain $\Omega$ into finite elements; (b) Linear hat functions illustrating local interpolation.}
  \label{fig:fem_overview}
\end{figure}

The finite element method (FEM) is a versatile numerical technique for solving boundary‐value problems governed by partial differential equations (PDEs). In FEM, the continuous domain $\Omega$ is discretized into a finite number of simple geometric subdomains called elements, each defined by a set of nodes $\{x_i\}_{i=1}^{n_e}$ and associated shape functions $\{N_i(x)\}_{i=1}^{n_e}$. Figure~\ref{fig:fem_overview} combines a 2D mesh illustration and a 1D hat‐function plot to clarify both the discretization process and the interpolation mechanism.

Within each element $e$, the unknown field $u(x)$ is interpolated by the nodal values $u_i$ using the shape functions
\begin{equation}
    u_h(x)\big|_{e} = \sum_{i=1}^{n_e} N_i(x)\,u_i,
\end{equation}
where $N_i(x)$ are typically low‐order polynomials (e.g., linear or quadratic) chosen to balance accuracy against computational cost.

Starting from the strong form of a governing PDE, for example, the linear elasticity equilibrium equation
\begin{equation}
  -\nabla\!\cdot\bigl(\mathbf{C}:\nabla u\bigr) = f \quad \text{in }\Omega,
\end{equation}
one derives the weak (variational) form by multiplying with a test function $v$, integrating over $\Omega$, and integrating by parts. Enforcing natural boundary conditions on $\Gamma_N$ yields
\begin{equation}
  \int_\Omega (\nabla v)^\mathsf{T}\,\mathbf{C}\,\nabla u_h\,\mathrm{d}\Omega
  = \int_\Omega v\,f\,\mathrm{d}\Omega
  + \int_{\Gamma_N} v\,\bar t\,\mathrm{d}\Gamma
  \quad \forall\,v\in V,
\end{equation}
where $V$ is the space of admissible test functions satisfying essential boundary conditions on $\Gamma_D$.

Each element contributes a local stiffness matrix $\mathbf{K}^e$ and load vector $\mathbf{F}^e$, which assemble into the global system
\begin{equation}
  \mathbf{K}\,\mathbf{U} = \mathbf{F}.
\end{equation}
Here, $\mathbf{K}\in\mathbb{R}^{N\times N}$ is a sparse, symmetric, positive‐definite matrix, $\mathbf{U}$ collects all nodal unknowns, and $\mathbf{F}$ is the global load vector. Efficient solution of this system typically employs direct solvers (e.g., LU or Cholesky factorization) or iterative solvers (e.g., conjugate gradient, generalized minimal residual method (GMRES) with preconditioning).

FEM excels at handling complex geometries, heterogeneous materials, and varied boundary conditions. Adaptive mesh refinement enhances accuracy by refining elements where solution gradients are high, while parallel solvers, multigrid techniques, and model‐order reduction mitigate computational cost.

\subsection{Spectral Methods}

Spectral methods solve boundary-value problems by approximating the unknown field $u(x)$ with a truncated global expansion
\begin{equation}
    u_N(x)=\sum_{k=0}^N \hat{u}_k\,\phi_k(x),
\end{equation}
where the basis functions $\{\phi_k\}$ are chosen to be orthogonal over the domain, such as Fourier modes on periodic intervals or Chebyshev polynomials on $[-1,1]$. Two main variants exist: the Galerkin spectral approach, which enforces orthogonality of the residual to the basis functions, and the pseudospectral (collocation) approach, which requires the PDE residual to vanish at a set of collocation points. By exploiting the global support of the basis, spectral methods achieve exponential (spectral) convergence for sufficiently smooth solutions: the error decays faster than any power of $N$. Computational efficiency is obtained via the fast Fourier transform for Fourier bases or via fast transforms for orthogonal polynomials.

Despite their high accuracy per degree of freedom, classical spectral methods are naturally restricted to simple geometries and uniform parameterizations. To extend spectral accuracy to complex domains, spectral element methods partition the domain into elements and apply high-order polynomial expansions within each element, combining the strengths of FEM and spectral techniques.

\subsection{Strengths and Limitations of Classical Methods}\label{sec:Strengths and Limitations of Classical Methods}

Classical numerical schemes for PDEs, including finite-difference, finite-element, finite-volume, and spectral methods, are underpinned by well-established theoretical foundations. Their convergence and stability properties are rigorously characterized, and reliable error estimates guide mesh refinement or basis‐order selection. A mature software ecosystem (for example, FEniCS, deal.II, and PETSc) offers production‐ready implementations, powerful meshing tools, and optimized solvers tailored for large‐scale sparse systems, making these methods the default choice in many engineering and physical applications.

However, several intrinsic challenges persist. The curse of dimensionality drives an exponential increase in degrees of freedom in high-dimensional problems, rapidly inflating computational cost and memory demands. Generating and adapting meshes becomes increasingly complex for evolving or highly irregular geometries, often requiring manual intervention or sophisticated algorithms. Furthermore, integrating observational data and solving inverse problems typically necessitates adjunct frameworks, such as adjoint-state methods, which add algorithmic and implementation complexity.

\section{Motivation for Neural‐Based Solvers}
The limitations of classical numerical methods for solving partial differential equations (discussed above) have driven the exploration of other approaches. Neural network-based solvers offer a paradigm shift by providing mesh-free approximations that can efficiently handle these complexities, motivated by the need for faster, more adaptable, and data-efficient solutions in fields like fluid dynamics, quantum mechanics, and biomedical engineering.

The first motivation came from the universal approximation theorem~\cite{hornik_multilayer_1989}, which posits that the neural networks can represent any continuous function to arbitrary accuracy, enabling them to parametrize PDE solutions directly without explicit discretization. This is good for high dimensional PDEs, where classical methods suffer exponential growth in computational cost. As deep learning technique and computing hardware advance, neural network solutions also improve. 

Another key driver is the ability to integrate physics and data addressing scenarios where traditional methods falter, such as inverse problems (e.g., parameter estimation from sparse measurements) or multi-physics simulations involving coupled equations. Neural solvers can learn from both simulated and real-world data, improving robustness and predictive accuracy for nonlinear or stiff PDEs.

In summary, the motivation for neural network-based PDE solvers lies in overcoming the bottlenecks of classical methods through flexibility, efficiency, and physics-data fusion, paving the way for broader applications in scientific computing and engineering.

\section{Primer on Neural Networks}
Feedforward neural networks (FNNs) provide the fundamental function approximation framework for PDE solvers, while automatic differentiation enables efficient and exact computation of the derivatives required by PDE residuals and variational formulations.

A feedforward neural network (FNN) defines a parametric mapping 
\begin{equation}
    \mathbf{y} = f_{\theta}(\mathbf{x}), \quad \theta = \{ W^{(l)}, \mathbf{b}^{(l)} \}_{l=1}^L,
\end{equation}
where $\mathbf{x} \in \mathbb{R}^{d_{\text{in}}}$ is the input vector, $\mathbf{y} \in \mathbb{R}^{d_{\text{out}}}$ is the output vector, and $\theta$ denotes all trainable weights and biases. The mapping is constructed as a sequence of affine transformations and nonlinear activations:
\begin{equation}
    \begin{aligned}
            \mathbf{h}^{(l)}& = \sigma\!\left( W^{(l)} \mathbf{h}^{(l-1)} + \mathbf{b}^{(l)} \right), 
    \quad l = 1, \dots, L-1,\\
    &\mathbf{y} = W^{(L)} \mathbf{h}^{(L-1)} + \mathbf{b}^{(L)},\\
    \end{aligned}
\end{equation}
with $\mathbf{h}^{(0)} = \mathbf{x}$ and $\sigma(\cdot)$ a nonlinear activation function (e.g., $\tanh$, $\sin$, or the rectified linear unit function, abbreviated ReLU).  
The universal approximation theorem~\cite{hornik_multilayer_1989} states that, given a sufficiently large number of neurons and a non-polynomial activation function, $f_{\theta}$ can approximate any continuous function on a compact domain to arbitrary accuracy. This makes FNNs natural candidates for representing the solutions $u(\mathbf{x})$ of PDEs, where $\mathbf{x}$ may include spatial and temporal coordinates.

A central requirement for neural network-based PDE solvers is the ability to evaluate derivatives of the network output $u_{\theta}(\mathbf{x})$ with respect to its input variables. For example, in a PDE of the form
\begin{equation}
    \mathcal{N}\big[u(\mathbf{x})\big] = 0, \quad \mathbf{x} \in \Omega,
\end{equation}
the operator $\mathcal{N}$ may involve first- or higher-order derivatives such as $\frac{\partial u}{\partial x_i}$ or $\frac{\partial^2 u}{\partial x_i \partial x_j}$. Computing these derivatives accurately is essential for forming the PDE residual.

Modern deep learning frameworks implement automatic differentiation (AD), which applies the chain rule of calculus to computational graphs. If the network output is given by a composition of elementary operations
\begin{equation}
    u_{\theta}(\mathbf{x}) = (f_L \circ f_{L-1} \circ \dots \circ f_1)(\mathbf{x}),
\end{equation}
then the gradient, with respect to either the parameters $\theta$ or the inputs $\mathbf{x}$, can be computed exactly (up to machine precision) by propagating derivatives backward through the graph:
\begin{equation}
        \frac{\partial u_{\theta}}{\partial \mathbf{x}} 
    = \frac{\partial f_L}{\partial f_{L-1}}
      \cdot \frac{\partial f_{L-1}}{\partial f_{L-2}}
      \cdots
      \frac{\partial f_1}{\partial \mathbf{x}}.
\end{equation}

This process, known as reverse-mode AD, has computational complexity comparable to a forward evaluation of the network, making it highly efficient.

For PDE learning, AD enables direct incorporation of the governing equations into the training loss. For instance, for the one-dimensional, time-independent Schrödinger equation
\begin{equation}
    -\frac{\hbar^2}{2m} \frac{\mathrm{d}^2 u}{\mathrm{d}x^2} + V(x) u(x) - E u(x) = 0,
\end{equation}
the second derivative $\frac{\mathrm{d}^2 u_{\theta}}{\mathrm{d}x^2}$ can be computed exactly via AD, allowing the residual
\begin{equation}
    \mathcal{R}(x) = -\frac{\hbar^2}{2m} u_{\theta}''(x) + V(x) u_{\theta}(x) - E u_{\theta}(x)
\end{equation}
to be embedded directly in the loss function. This capability eliminates the need for symbolic differentiation or numerical finite differences, both of which can be cumbersome or inaccurate for high-dimensional problems. However, it requires the activation function to be sufficiently differentiable. For example, ReLU would not be suitable for second-order PDEs due to its non-differentiability at zero.\SA{Mention that this requires the activation function to be "differentiable enough". For example ReLu would not work for 2nd order PDEs.)}

In summary, feedforward neural networks offer a flexible function representation, and automatic differentiation provides an exact and efficient means of evaluating the derivatives required by PDE residuals or variational forms. Together, these properties form the computational backbone of modern neural-network-based PDE solvers such as physics-informed neural networks (PINNs), the deep Ritz method (DRM), and weak adversarial networks (WANs).

\section{A Broad Survey of Neural PDE Methods}
The idea of representing the solution to partial differential equations (PDEs) using neural network trial functions can be traced back to the 1990s, when Dissanayake and Phan-Thien first proposed a neural trial-function approach for PDEs. However, due to limitations in computational hardware and neural network theory at the time, these methods were not widely adopted. The advent of modern deep learning, particularly the development of automatic differentiation frameworks, revolutionized this landscape by enabling the efficient computation of PDE residuals directly within neural network training pipelines. This breakthrough made it possible to embed the governing equations of physical systems directly into the loss function, leading to the rapid development of neural network-based PDE solvers.

In the following, we review recent neural network-based methods for solving PDEs, focusing on three major frameworks, physics-informed neural networks (PINNs), the deep Ritz method (DRM), and weak adversarial networks (WANs), before highlighting several other significant developments in the past five years.

\subsection{Physics-Informed Neural Networks}

PINNs have emerged as a transformative approach in scientific computing, offering a mesh-free and flexible framework for solving both forward and inverse PDE problems. Introduced in their modern form by Raissi, Perdikaris, and Karniadakis~\cite{raissi_physics-informed_2019}, PINNs employ a deep feedforward neural network trained to minimize a composite loss function that includes the PDE residual and any relevant boundary or initial condition errors. This ensures that the learned surrogate model not only fits observational data (if available) but also strictly adheres to the governing physical laws.

The general PINN workflow involves sampling collocation points in the domain, evaluating the PDE residual at these points via automatic differentiation, and adjusting the network weights to minimize residuals and boundary mismatches. The same framework extends naturally to inverse problems by treating unknown parameters as additional trainable variables. PINNs have been successfully applied to diverse fields including fluid dynamics, solid mechanics, and heat transfer, often achieving high accuracy with sparse data.

Despite their versatility, vanilla PINNs can suffer from slow or unstable convergence, particularly for stiff PDEs or when the loss terms (PDE residuals vs. boundary conditions) have competing scales. In response, several extensions have been proposed. One line of work focuses on domain decomposition and parallelization, exemplified by extended PINNs (XPINNs)~\cite{jagtap_extended_2020}, which generalize PINNs to a space–time domain decomposition using multiple subnetworks trained in parallel on smaller subdomains to reduce learning complexity and improve scalability. Another set of improvements targets adaptive sampling and loss balancing, including methods that dynamically adjust collocation points and loss weights to stabilize training. Probabilistic formulations, such as Bayesian PINNs (BPINNs)~\cite{yang_b-pinns_2020}, incorporate Bayesian inference to provide uncertainty estimates over the predicted solution. Variational PINNs (VPINNs)~\cite{kharazmi_variational_2019}, reformulate the PDE in variational form, minimizing an energy functional rather than the raw residual. Architectural variants have also been developed, including physics-informed convolutional and recurrent networks for spatio-temporal PDEs. Hybrid pre-processing approaches, such as the physics-informed Gaussian (PIG) model~\cite{kang_pig_2025}, combine feature embeddings using Gaussian functions with PINN, improving accuracy in data-sparse regions.

\subsection{The Deep Ritz Method}

When a PDE admits a variational (energy) formulation, the deep Ritz method introduced by E and Yu~\cite{e_deep_2017} offers a compelling alternative. Instead of enforcing the PDE’s differential form, the deep Ritz method minimizes an energy functional whose minimizer is the PDE solution. This approach naturally accommodates nonlinear PDEs, adapts to regions of high solution complexity, and scales well to high-dimensional problems using Monte Carlo integration for functional evaluation. The variational framework has inspired further developments, such as the deep double Ritz method~\cite{uriarte_deep_2023}, which introduces a nested optimization in which one network approximates the trial solution, while the other optimizes the test functions in an inner loop, enhancing accuracy for complex variational problems.

\subsection{Weak Adversarial Networks}

Weak adversarial networks, proposed by Zang et al.~\cite{zang_weak_2020}, reformulate PDEs in their weak form and solve them via adversarial optimization. Two networks are trained simultaneously: a primal network \( u_\theta(x) \) approximates the PDE solution, while an adversarial test network \( v_\phi(x) \) seeks functions that maximize the residual of the weak form. This setup creates a min-max game analogous to generative adversarial networks (GANs), driving the solution network to satisfy the PDE for all admissible test functions. WANs are particularly well-suited for PDEs where strong-form residuals are difficult to compute or unstable.

\subsection{Other Neural Network PDE Solvers}

Several other architectures expand beyond these three frameworks. The deep Galerkin method (DGM), introduced by Sirignano and Spiliopoulos~\cite{sirignano_dgm_2018}, targets high-dimensional PDEs (up to 100–200 dimensions) by directly sampling and minimizing residuals without a mesh. Another rapidly growing area is operator learning, where the goal is to learn the entire PDE solution operator from data. Notable examples include DeepONet~\cite{lu_deeponet_2021} and the Fourier neural operator (FNO)~\cite{li_fourier_2021}, the latter operating in the frequency domain to efficiently learn solution mappings for families of PDEs.

\subsection{Applications to Schrödinger Equations}

Neural network-based PDE solvers have found significant application in wave dynamics and quantum systems, particularly involving Schrödinger-type equations. Pu \textit{et al.}~\cite{pu_solving_2021} developed an enhanced PINN with neuron-wise locally adaptive activation functions for the derivative nonlinear Schrödinger equation (DNLS), achieving accelerated convergence and improved fidelity in simulating soliton and rogue wave solutions. Wang \textit{et al.}~\cite{wang_data-driven_2021} extended multi-layer PINNs to recover data-driven rogue wave solutions of the defocusing NLS equation, demonstrating the method’s ability to capture parameter-sensitive nonlinear phenomena. Yuan \textit{et al.}~\cite{yuan_physics-informed_2023} applied PINNs to modified NLS models, highlighting their adaptability to varying governing equations. Harcombe \textit{et al.}~\cite{harcombe_physics-informed_2023} further showcased PINNs in identifying localized eigenstates in disordered media, extending their utility beyond conventional wave systems.

Collectively, these studies underscore the rapid evolution of neural network-based PDE solvers, from foundational PINNs to variational and adversarial formulations, and from general-purpose frameworks to specialized adaptations for complex wave phenomena, marking a vibrant and expanding frontier in computational physics and applied mathematics.

\section{Summary}

In this chapter, we have reviewed the fundamental concepts of partial differential equations (PDEs) and neural networks, followed by an overview of classical numerical methods for solving PDEs and the emerging class of neural-network-based approaches. Particular emphasis was placed on three prominent frameworks: physics-informed neural networks (PINNs), the deep Ritz method (DRM), and weak adversarial networks (WANs), along with their recent developments and applications.

In the subsequent chapters, we will systematically compare the performance of PINNs, DRM, and WANs to both the Schrödinger equation and selected benchmark problems, with a focus on predicting higher-order wavefunctions and the treatment of higher-dimensional cases. Detailed mathematical descriptions of these methods are provided in Chapter 3. Finally, we will apply PINNs, DRM, and WANs to a novel potential in strong-field laser physics, the Kramers-Henneberger (KH) potential, and assess their performance in this context.

\SA{Good overview chapter. Because you left out any mathematics when describing the methods (PINN, DRM, WAN) you could refer forwards to the sections in chapter 3 where you will have those details.}

\chapter{Methodology}
This chapter presents the methodological framework used to implement and compare three neural network based solvers for partial differential equations (PDEs): physics informed neural networks (PINNs), weak adversarial networks (WANs), and the deep Ritz method (DRM). The study focuses on two canonical PDE families, the Poisson equation and the time independent Schr\"odinger equation with representative potentials (infinite well and harmonic oscillator). The goal is to establish a common basis for fair comparison of accuracy, stability, and efficiency.

We first formalize each problem class by specifying the governing equation, the spatial domain \(\Omega\), and the boundary conditions. Where analytic references exist, we provide exact solutions to anchor quantitative evaluation. We then introduce the three solvers in a unified notation, derive their training objectives, and outline the general workflow for each method, including sampling and constraint handling. Finally, we describe the implementation and evaluation protocol, covering the computing platform and software stack, the network architectures and training schedules, the hyperparameters used across tasks, and the metrics reported, such as the relative \(L_2\) error on holdout grids, boundary residuals, and wall clock time. 

\section{Problem Setups}\label{problem setups}
To fully examine the capabilities of different neural network–based methods, we select a set of benchmark PDEs that are both well understood and representative of common challenges. We focus on the Poisson equation as a canonical elliptic problem, and on the time‑independent Schrödinger equation with several potentials to test eigenvalue and normalization constraints. Wherever possible, exact solutions are provided for quantitative comparison.

\subsection{Poisson's Equation}

Poisson's equation is a canonical linear elliptic PDE that arises in electrostatics, Newtonian gravitation, and many other physical settings. Its general form, in $d$ spatial dimensions, is
\begin{equation}
  -\Delta u(\mathbf{x}) = f(\mathbf{x}), \qquad \mathbf{x}\in\Omega\subset\mathbb{R}^d,
\end{equation}
where $u:\Omega\to\mathbb{R}$ is the unknown field, $f:\Omega\to\mathbb{R}$ is a prescribed source term, and $\Omega$ is a bounded domain with boundary $\partial\Omega$. In this work, we focus on homogeneous Dirichlet boundary conditions:
\begin{equation}
  u(\mathbf{x}) = 0, \quad \mathbf{x}\in\partial\Omega.
\end{equation}

As an analytic benchmark, let $\Omega=(0,1)^d$ and choose
\begin{equation}
  f(\mathbf{x}) = \prod_{i=1}^d \sin(\pi x_i), \qquad \mathbf{x}=(x_1,\dots,x_d).
\end{equation}
Then the boundary-value problem
\begin{equation}
\begin{cases}
  -\Delta u(\mathbf{x}) = \displaystyle \prod_{i=1}^d \sin(\pi x_i), & \mathbf{x}\in(0,1)^d,\\[6pt]
  u(\mathbf{x}) = 0, & \mathbf{x}\in\partial(0,1)^d,
\end{cases}
\end{equation}
has the exact solution
\begin{equation}
  u(\mathbf{x}) = \frac{1}{d\,\pi^2}\,\prod_{i=1}^d \sin(\pi x_i).
\end{equation}
Indeed, since $\Delta\!\left(\prod_{i=1}^d \sin(\pi x_i)\right) = -d\,\pi^2 \prod_{i=1}^d \sin(\pi x_i)$ and the product vanishes on $\partial(0,1)^d$, the above $u$ satisfies both the PDE and the boundary conditions. The familiar one- and two-dimensional cases are recovered by setting $d=1$ and $d=2$:
\begin{equation}
\begin{aligned}
    u(x)& = \frac{1}{\pi^2}\sin(\pi x),\\
    u(x,y)& = \frac{1}{2\pi^2}\sin(\pi x)\sin(\pi y).\\
\end{aligned}
\end{equation}

If $\Omega=(0,L)^d$, an equally convenient choice is
\begin{equation}
  f(\mathbf{x}) = \prod_{i=1}^d \sin\!\left(\frac{\pi x_i}{L}\right),
\end{equation}
for which the unique solution with $u|_{\partial\Omega}=0$ is
\begin{equation}
  u(\mathbf{x}) = \frac{L^2}{d\,\pi^2}\,\prod_{i=1}^d \sin\!\left(\frac{\pi x_i}{L}\right).
\end{equation}
This construction extends naturally to any spatial dimensionality $d\in\mathbb{N}$. In our experiments, we will investigate from $d=1$ to $d=5$.

\subsection{Time-Independent Schr\"odinger equation}

The time-independent Schr\"odinger equation governs stationary states of nonrelativistic quantum systems and is widely used across physics and chemistry. In $d$ spatial dimensions on a domain $\Omega\subset\mathbb{R}^d$, it reads
\begin{equation}
  -\frac{\hbar^2}{2m}\,\Delta \psi(\mathbf{x}) + V(\mathbf{x})\,\psi(\mathbf{x})
  = E\,\psi(\mathbf{x}), \qquad \mathbf{x}\in\Omega,
\end{equation}
where $\psi(\mathbf{x})$ is the wavefunction, $V(\mathbf{x})$ the external potential, and $E$ the associated energy. For bound states, the Born interpretation requires normalization:
\begin{equation}
  \int_{\Omega} |\psi(\mathbf{x})|^2\, d\mathbf{x} = 1,
\end{equation}
while boundary conditions are dictated by the physics of the problem. For example, $\psi(\mathbf{x})\to 0$ as $\|\mathbf{x}\|\to\infty$ on unbounded domains, or homogeneous Dirichlet conditions on impenetrable walls.

In what follows, we compute ground and excited stationary states for two canonical benchmark potentials: the infinite square well and the harmonic oscillator.

\subsubsection{Infinite Potential Well}

\begin{figure}[ht]
  \centering
  \begin{subfigure}[b]{0.45\textwidth}
    \centering
    \includegraphics[height=3cm,keepaspectratio]{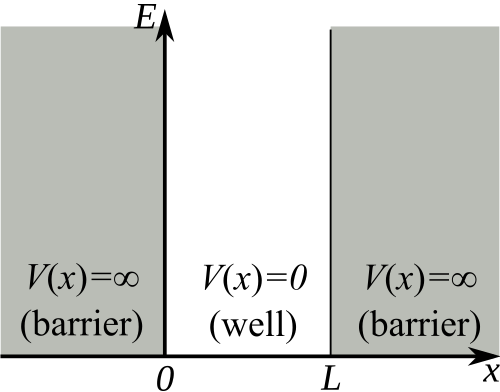}
    \caption{Infinite square-well potential in 1D.}
    \label{fig:well1d}
  \end{subfigure}
  \quad
  \begin{subfigure}[b]{0.45\textwidth}
    \centering
    \includegraphics[height=3cm,keepaspectratio]{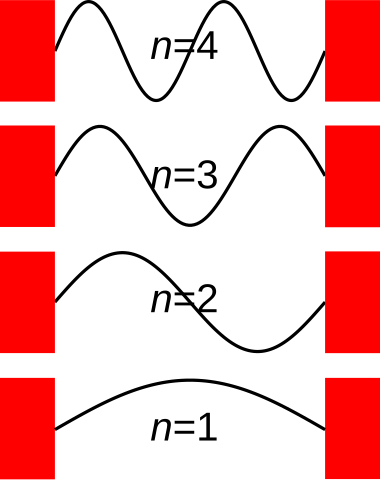}
    \caption{wavefunctions $\psi_n(x)$.}
    \label{fig:prob}
  \end{subfigure}
  \caption{Infinite square well and representative stationary states~\cite{papa_november_particle_2025}.}
  
  \label{1d_se_plot}
\end{figure}

The infinite potential well is an archetypal and exactly solvable model in quantum mechanics: a particle is confined to a finite region with zero potential inside and an infinite barrier outside (see Figure \ref{1d_se_plot}). The wavefunction, therefore, satisfies homogeneous Dirichlet boundary conditions (i.e. it vanishes on the walls). Because closed-form solutions are available, the model is ideal for validating numerical methods while isolating fundamental effects such as energy quantization.

In one dimension, on the interval $\Omega=(0,L)$, the time-independent Schr\"odinger equation with boundary conditions is
\begin{equation}
\begin{cases}
-\dfrac{\hbar^2}{2m}\,\dfrac{d^2\psi}{dx^2} = E\,\psi, & x\in(0,L),\\[6pt]
\psi(0)=\psi(L)=0.
\end{cases}
\end{equation}
The normalized stationary states and their energies are
\begin{equation}
\psi_n(x)=\sqrt{\frac{2}{L}}\sin\!\left(\frac{n\pi x}{L}\right),
\quad
E_n=\frac{\hbar^2\pi^2}{2mL^2}\,n^2,
\quad n=1,2,\ldots.
\end{equation}

In two dimensions, for a rectangular box $\Omega=(0,L_x)\times(0,L_y)$, the problem reads
\begin{equation}
\begin{cases}
-\dfrac{\hbar^2}{2m}\,\big(\partial_{xx}+\partial_{yy}\big)\,\psi(x,y)=E\,\psi(x,y), & (x,y)\in\Omega,\\[6pt]
\psi=0, & \text{on }\partial\Omega.
\end{cases}
\end{equation}
Following separation of variables, the normalized solutions and corresponding energies are
\begin{equation}
\begin{aligned}
\psi_{n_x,n_y}(x,y)
&= \sqrt{\frac{2}{L_x}}\,\sqrt{\frac{2}{L_y}}\,
   \sin\!\left(\frac{n_x\pi x}{L_x}\right)\,
   \sin\!\left(\frac{n_y\pi y}{L_y}\right),\\[6pt]
E_{n_x,n_y}
&= \frac{\hbar^2\pi^2}{2m}\!\left(\frac{n_x^2}{L_x^2}+\frac{n_y^2}{L_y^2}\right),
\quad n_x,n_y\in\mathbb{N}.
\end{aligned}
\end{equation}

Degeneracy arises naturally in two dimensions. For a square box ($L_x=L_y$), the spectrum depends only on $n_x^2+n_y^2$; for example, $(n_x,n_y)=(1,2)$ and $(2,1)$ share the same energy.

\subsubsection{Quantum Harmonic Oscillator}

\begin{figure}[ht]
  \centering
  \begin{subfigure}[b]{0.45\textwidth}
    \centering
    \includegraphics[height=5cm,keepaspectratio]{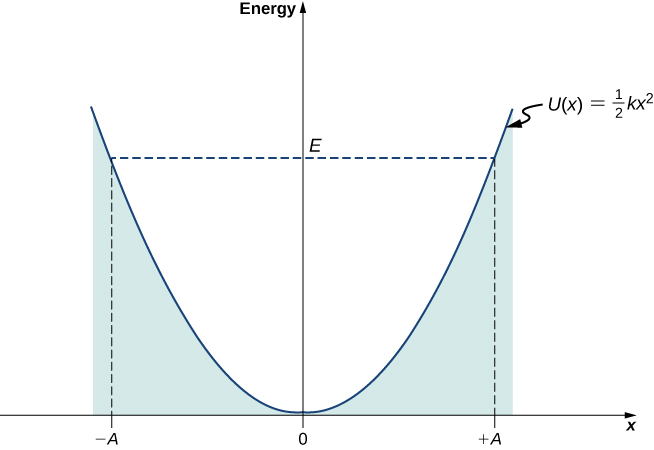}
    \caption{Harmonic potential in 1D.}
    \label{fig:well1d_qho}
  \end{subfigure}
  \quad
  \begin{subfigure}[b]{0.45\textwidth}
    \centering
    \includegraphics[height=5cm,keepaspectratio]{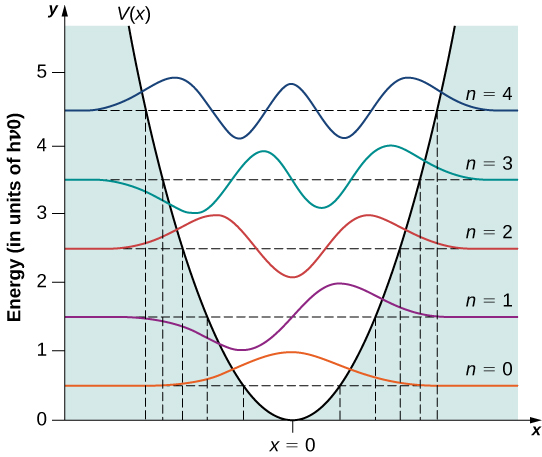}
    \caption{Wavefunctions $\psi_n(x)$.}
    \label{fig:prob_qhoS}
  \end{subfigure}
  \caption{Harmonic oscillator potential and representative stationary states~\cite{ling_university_nodate}.}
  \label{1d_qho_plot}
\end{figure}

The quantum harmonic oscillator is the canonical quantum analogue of the classical spring-mass system. It models, among many applications, the vibrational motion of atoms in crystals and molecular normal modes. The confining parabolic potential drives the wavefunction to decay at infinity, and closed-form solutions make the problem a reliable benchmark for validating numerical methods. A representative one-dimensional case, together with its potential and stationary states, is shown in Figure~\ref{1d_qho_plot}.

In one dimension $(x\in\mathbb{R})$, the time-independent Schr\"odinger equation is
\begin{equation}
  -\frac{\hbar^2}{2m}\,\frac{d^2\psi}{dx^2} + \frac{1}{2}m\omega^2 x^2\,\psi(x) = E\,\psi(x),
\end{equation}
with normalizable solutions
\begin{equation}
  \psi_n(x)
  = \frac{1}{\sqrt{2^n n!}}\left(\frac{m\omega}{\pi\hbar}\right)^{\!1/4}
    e^{-\frac{m\omega x^2}{2\hbar}}
    H_n\!\left(\sqrt{\frac{m\omega}{\hbar}}\,x\right),
  \quad n=0,1,2,\ldots
\end{equation}
and equally spaced energies
\begin{equation}
  E_n = \hbar\omega\!\left(n+\tfrac{1}{2}\right).
\end{equation}
Here, $H_n$ are the physicists' Hermite polynomials,
\begin{equation}
  H_n(z) = (-1)^n e^{z^2}\,\frac{d^n}{dz^n}\!\left(e^{-z^2}\right).
\end{equation}
Note that $n$ starts at $0$, and the ground-state energy $E_0=\tfrac{1}{2}\hbar\omega$ reflects the zero-point motion mandated by the uncertainty principle.

In two dimensions $(x,y\in\mathbb{R})$, the Hamiltonian separates. For a (possibly anisotropic) oscillator with frequencies $\omega_x,\omega_y>0$,
\begin{equation}
  -\frac{\hbar^2}{2m}\,(\partial_{xx}+\partial_{yy})\,\psi(x,y)
  + \frac{1}{2}m\!\left(\omega_x^2 x^2 + \omega_y^2 y^2\right)\psi(x,y)
  = E\,\psi(x,y).
\end{equation}
Separable, normalized stationary states and their energies are
\begin{equation}
\begin{aligned}
\psi_{n_x,n_y}(x,y)
&= \frac{1}{\sqrt{2^{\,n_x+n_y}\,n_x!\,n_y!}}
   \left(\frac{m\sqrt{\omega_x\omega_y}}{\pi\hbar}\right)^{\!1/2}
   \exp\!\left[-\frac{m}{2\hbar}\!\left(\omega_x x^2+\omega_y y^2\right)\right]\\
&\quad\times
   H_{n_x}\!\left(\sqrt{\frac{m\omega_x}{\hbar}}\,x\right)\,
   H_{n_y}\!\left(\sqrt{\frac{m\omega_y}{\hbar}}\,y\right),\\[6pt]
E_{n_x,n_y}
&= \hbar\omega_x\!\left(n_x+\tfrac{1}{2}\right)
 + \hbar\omega_y\!\left(n_y+\tfrac{1}{2}\right),
\quad n_x,n_y\in\mathbb{N}_0.
\end{aligned}
\end{equation}
In the isotropic case ($\omega_x=\omega_y=\omega$), the spectrum depends only on $N=n_x+n_y$:
\begin{equation}
  E_N = \hbar\omega\,(N+1), \quad N=0,1,2,\ldots,
\end{equation}
with degeneracy $N+1$ arising from all pairs $(n_x,n_y)$ such that $n_x+n_y=N$.

\section{Neural Network-Based Methods}
In following parts, we introduce three neural network based approaches for solving PDEs: physics–informed neural networks (PINNs), the deep Ritz method (DRM), and weak adversarial networks (WANs). Each method employs a neural network Ansatz to approximate the solution while incorporating the underlying physical or variational structure of the PDE.

\subsection{Physics–Informed Neural Networks}
PINNs \cite{raissi_physics-informed_2019} embed the governing equations directly into the loss function. Let
\begin{equation}
\begin{cases}
\mathcal{N}[u](\mathbf{x}) = 0, & \mathbf{x}\in\Omega,\\
u(\mathbf{x}) = g(\mathbf{x}), & \mathbf{x}\in\partial\Omega,
\end{cases}
\end{equation}
denote a general PDE with boundary data \(g\). We approximate \(u\) by a neural network \(u(\mathbf{x};\theta)\). Define three loss components:
\begin{equation}
\mathcal{L}_{\rm int}(\theta)
= \frac{1}{N_{\rm int}}\sum_{i=1}^{N_{\rm int}}
\bigl|\mathcal{N}[u(\mathbf{x}_i^{\rm int};\theta)]\bigr|^2,
\end{equation}
\begin{equation}
\mathcal{L}_{\rm bc}(\theta)
= \frac{1}{N_{\rm bc}}\sum_{i=1}^{N_{\rm bc}}
\bigl|u(\mathbf{x}_i^{\rm bc};\theta)-g(\mathbf{x}_i^{\rm bc})\bigr|^2,
\end{equation}
\begin{equation}
\mathcal{L}_{\rm data}(\theta)
= \frac{1}{N_{d}}\sum_{i=1}^{N_{d}}
\bigl|u(\mathbf{x}_i^{d};\theta)-g(\mathbf{x}_i^{d})\bigr|^2,
\end{equation}
where \(N_{\rm int},N_{\rm bc},N_{d}\) are the counts of the interior, boundary, and (optional) measurement points, respectively. The total loss is a weighted sum,
\begin{equation}
\mathcal{L}(\theta)
= \lambda_{\rm int}\,\mathcal{L}_{\rm int}
+ \lambda_{\rm bc}\,\mathcal{L}_{\rm bc}
+ \lambda_{d}\,\mathcal{L}_{\rm data}.
\end{equation}
Tuning \(\{\lambda\}\) involves balancing the enforcement of the PDE residual, boundary conditions, and any available data. Training proceeds by sampling collocation points in \(\Omega\) and \(\partial\Omega\), computing gradients via automatic differentiation, and optimizing \(\theta\) with Adam~\cite{kingma_adam_2017} or L–BFGS~\cite{liu_limited_1989}.  
\begin{figure}[ht]
    \centering
    \includegraphics[width=0.99\textwidth]{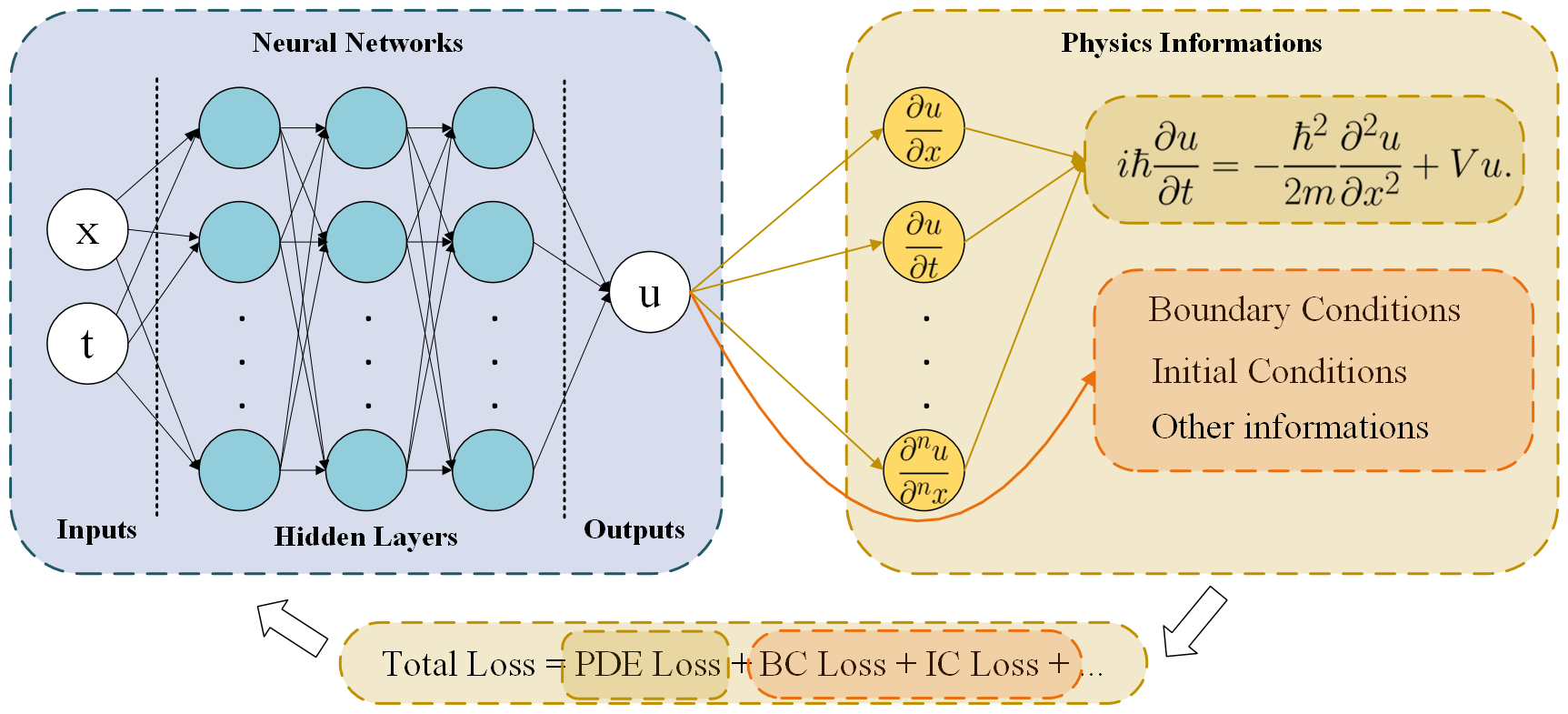}
    \caption{Workflow of a PINN solving the time–dependent Schrödinger equation.}
    \label{fig:PINN-workflow}
\end{figure}

\subsection{Deep Ritz Method}
The deep Ritz method \cite{e_deep_2017} reformulates a self–adjoint PDE as the minimization of an energy functional. For example, the Poisson problem
\begin{equation}
-\Delta u = f\text{ in }\Omega,\quad u=0\text{ on }\partial\Omega
\end{equation}
has the variational form
\begin{equation}
\min_{u\in H_0^1(\Omega)}
I[u]
\text{ with }
I[u]
=\int_\Omega\!\Bigl(\tfrac12|\nabla u|^2 - f\,u\Bigr)\,d\mathbf{x}.
\end{equation}
We approximate \(u(\mathbf{x};\theta)\) by a neural network and replace the integral by Monte Carlo sampling:
\begin{equation}
I_N(\theta)
=\frac{|\Omega|}{N}\sum_{i=1}^N\Bigl(\tfrac12|\nabla u(\mathbf{x}_i;\theta)|^2 - f(\mathbf{x}_i)\,u(\mathbf{x}_i;\theta)\Bigr).
\end{equation}
Dirichlet conditions are enforced by adding a penalty term:
\begin{equation}
\widetilde I_N(\theta)
=I_N(\theta)
+\beta\;\frac{|\partial\Omega|}{N_{\partial}}
\sum_{j=1}^{N_{\partial}}
\bigl|u(\mathbf{x}_j^{\partial};\theta)\bigr|^2,
\end{equation}
with large \(\beta\). Minimizing \(\widetilde I_N\) recovers the weak solution.  

For eigenvalue problems such as the time–independent Schrödinger equation, one minimizes the Rayleigh quotient,
\begin{equation}
E[\psi]
=\frac{\displaystyle\int\Bigl(\frac{\hbar^2}{2m}|\nabla\psi|^2 + V|\psi|^2\Bigr)\,d\mathbf{x}}
{\displaystyle\int|\psi|^2\,d\mathbf{x}},
\end{equation}
subject to \(\|\psi\|_{L^2}=1\). In practice, one optimizes
\begin{equation}
\mathcal{L}(\theta)
=J_N(\theta)
+ \lambda_{\rm norm}\Bigl(\|\psi(\cdot;\theta)\|^2_{L^2}-1\Bigr)^2
+ \beta\times\text{BC penalty},
\end{equation}
where \(J_N\) is the Monte Carlo-approximated numerator. Higher eigenstates impose additional orthogonality penalties \(\int\psi\,\psi_k\,d\mathbf{x}=0\).  

\subsection{Weak Adversarial Networks}
\begin{figure}[ht]
  \centering
  \includegraphics[width=0.99\textwidth]{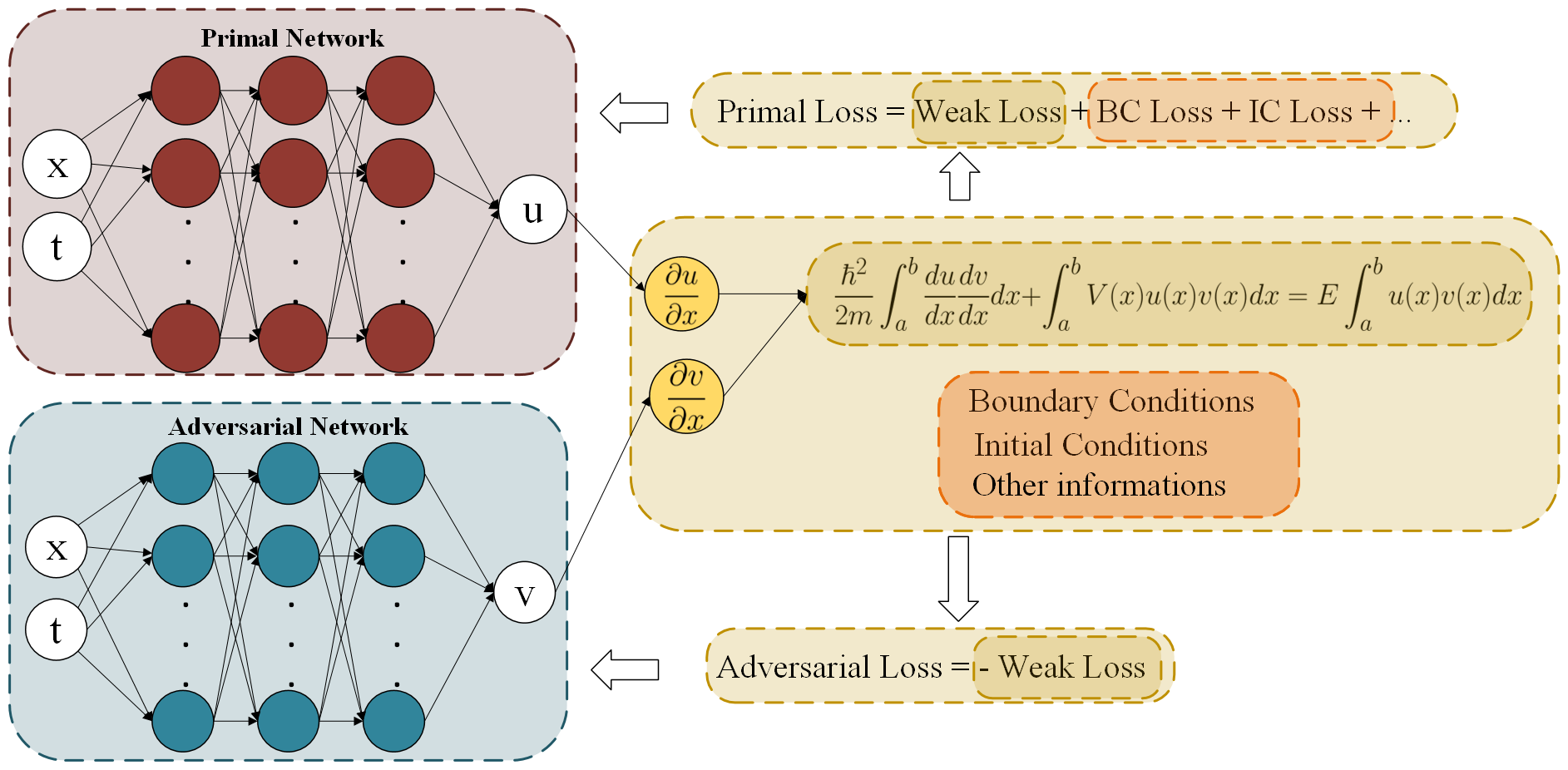}
  \caption{WAN architecture applied to the Schrödinger equation in its weak form.}
  \label{fig:WAN-workflow}
\end{figure}

WANs \cite{zang_weak_2020}, illustrated in Figure \ref{fig:WAN-workflow}, blend the weak formulation with an adversarial training paradigm. Starting from the bilinear form for a general PDE,
\begin{equation}
\langle\mathcal{A}[u],v\rangle
=\int_\Omega\bigl(\nabla u\cdot\nabla v - f\,v\bigr)\,d\mathbf{x},
\end{equation}
A WAN combines two networks:
\begin{itemize}
  \item a primal network \(u(\mathbf{x};\theta)\) approximating the solution, and
  \item an adversarial network \(v(\mathbf{x};\eta)\) serving as a test function.
\end{itemize}
The weak residual is
\begin{equation}
\mathcal{R}(\theta,\eta)
=\bigl|\langle\mathcal{A}[u(\cdot;\theta)],v(\cdot;\eta)\rangle\bigr|^2.
\end{equation}
Training solves the saddle–point problem
\begin{equation}
\min_\theta\max_\eta\;
\lambda_{\rm weak}\,\mathcal{R}(\theta,\eta)
+ \lambda_{\rm bc}\,\mathcal{L}_{\rm bc}(\theta),
\end{equation}
where \(\mathcal{L}_{\rm bc}\) enforces boundary conditions as before. In each iteration, \(\eta\) is updated to maximize the weak residual (identifying the worst-case test function), and \(\theta\) is updated to minimize it. This adversarial coupling enhances stability in high dimensions and complex geometries.  

\section{Implementation Details and Evaluation Protocol}

\subsection{Experimental Platform and Software}

All experiments were conducted on a laptop equipped with an NVIDIA GeForce RTX 3080 GPU (running CUDA version 11.7) and an Intel Core i9 CPU, supported by 32 GB of RAM. The software environment consisted of Python 3.9, PyTorch 1.12, and NumPy 1.22. To ensure reproducibility, random seeds for PyTorch, NumPy, and CUDA were fixed at runtime. The complete codebase, including the \texttt{requirement.txt} specification and scripts for data generation, training, and evaluation, is publicly available at \url{https://github.com/JiakangC/Neural-Network-Based-PDE-Solver.git}.

\subsection{Model Architectures and Training}

We evaluated a fully connected network (FCN) as our baseline model, comprising three hidden layers with 50 neurons each and employing hyperbolic tangent activation functions. All weights were initialized using Xavier initialization to promote stable training dynamics. Optimization was performed using the Adam optimizer with an initial learning rate of $10^{-3}$, over a maximum of \num{10000} to \num{50000} epochs.

The loss function consisted of at least three primary terms: the PDE or weak-form residual, the boundary condition enforcement, and an optional data supervision term. The coefficients for these terms were carefully tuned within the range $[1, \num{10000}]$ to balance their contributions effectively. Although the supervised data term is not essential for PINNs, our experiments demonstrated that incorporating even a modest amount of labelled data significantly enhanced solution accuracy. To systematically explore this effect, we trained the network on the first 25\% of the available data.

For enhanced performance, we further explored architectural modifications, including increasing the network depth (e.g., adding more hidden layers) and the number of neurons per layer. Additionally, we introduced custom sine activation functions, which have been shown to better capture oscillatory behaviours inherent in certain PDE solutions, such as those from quantum mechanics. To further strengthen the physics-informed aspects of the model, we incorporated additional loss terms designed to enforce specific physical constraints; these will be detailed below.

\subsection{Evaluation Protocol}

Model performance was primarily quantified using the relative $L_2$ error, defined as
\begin{equation}
\varepsilon_{L_2} = \frac{\|u_{\mathrm{NN}} - u_{\mathrm{exact}}\|_{L_2(\Omega)}}{\|u_{\mathrm{exact}}\|_{L_2(\Omega)}},
\end{equation}
along with the training time required to achieve convergence. For problems involving the Schrödinger equation, we also report the eigenvalue error $|E_{\mathrm{NN}} - E_{\mathrm{exact}}|$, where applicable, to assess the accuracy of energy level predictions. Throughout the training process, the PDE residual norm was monitored to track convergence and ensure adherence to the governing equations.

Computational efficiency was characterized by recording the runtime to reach the best $L_2$ error and average duration per epoch, offering insights into the practical costs of the methods.

\subsection{Numerical Experiments}

We applied the proposed framework to the benchmark problems outlined in Section~\ref{problem setups}, beginning with Poisson equation tests, from one to five dimensions, to validate the basic setup, before advancing to quantum mechanical systems such as the infinite square well and the quantum harmonic oscillator. Our primary focus, however, was on the Schrödinger equations, where the physics-informed approaches demonstrate particular advantages in handling eigenvalue problems and boundary conditions.

As an initial testbed, we considered one-dimensional problems. For the one-dimensional infinite potential well, we adopted the foundational approach from Raissi et al.~\cite{raissi_physics-informed_2019}, balancing the PDE residual, boundary conditions, and optional data terms via tunable weight parameters, as described in Section 3.2. This configuration served as our baseline and was denoted as ``BC" (Boundary Condition weighting). To enhance boundary enforcement, we introduced a forced boundary condition directly into the neural network architecture; we label this variant ``FB" (Forced Boundary). Recognizing the need for consistency across multiple energy states in eigenvalue problems, we proposed two extensions to handle higher modes effectively.

The first extension incorporated an orthogonality penalty, a technique commonly used to ensure that predicted eigenfunctions remain orthogonal, as required by quantum mechanics; we refer to this extension as ``OG" (Orthogonality). The second novel approach, termed ``FN" (Forced Nodes), involved strategically controlling the positions of nodal points (where the wave function vanishes) to partition the domain into smaller subregions with enforced zero boundaries. This decomposition simplifies the learning task by breaking it into more manageable segments. Detailed descriptions of these techniques, including their implementation and rationale, are presented alongside the results in the next chapter.

The same suite of techniques, BC, FB, OG, and FN, was subsequently applied to the one-dimensional quantum harmonic oscillator. In this problem, we introduced an additional trainable parameter for the energy eigenvalue, initialized with the exact value as a guess to guide the optimization process and improve convergence.

Finally, we extended these methodologies to higher-dimensional settings. Using FB as the baseline for boundary enforcement, we compared it against OG and FN for both the infinite potential well and the quantum harmonic oscillator in two dimensions. This progression allowed us to evaluate the scalability of the approaches, assessing how well they generalize to increased complexity while maintaining accuracy and efficiency.

\chapter{Results and Analysis for Standard Problems}
This chapter presents the empirical findings from applying physics-informed neural networks (PINNs), deep Ritz method (DRM) and weak adversarial networks (WANs) to the baseline Poisson equation and standard Schrödinger problems (infinite potential well in 1D and 2D; quantum harmonic oscillator in 1D and 2D). To provide a comprehensive evaluation, results are divided into baseline implementations (standard setups without modifications) and optimized versions (incorporating improvements like additional loss terms, hard boundary conditions, and advanced architectures). This structure highlights initial limitations (e.g., high$L_2$errors in baselines due to trivial solutions or spectral bias) and demonstrates how targeted enhancements improve performance, enabling fair cross-method comparisons. 

\section{Baseline Results}
\subsection{Poisson Equation Baselines}
We evaluate the efficacy of our method in solving the Poisson equation across multiple dimensions without relying on any training data. In the following table, we present the performance from one to five dimensions, considering two distinct boundary conditions: `BC' for weight-controlled boundaries and `FB' for forced boundaries. Table~\ref{Results for Multi-dimensional Poisson Equation} summarizes these outcomes.
\begin{table}[ht]
\centering
\caption{Results for Multi-dimensional Poisson Equation}
\label{Results for Multi-dimensional Poisson Equation}
\begin{tabular}{l l  c c c c c }
\toprule
& \text{Dim}& $1$ & $2$ & $3$ & $4$ & $5$ \\
\midrule
\multirow{2}{*}{PINN} & BC & $2.58\textsc{e}{-}03$
 & $1.87\textsc{e}{-}02$
 & $2.94\textsc{e}{-}02$
 & $3.75\textsc{e}{-}02$
 & $1.29\textsc{e}{-}01$
 \\
& FB & \cellcolor[gray]{.8}$4.11\textsc{e}{-}05$
 & \cellcolor[gray]{.8}$3.20\textsc{e}{-}05$
 & \cellcolor[gray]{.8}$6.27\textsc{e}{-}05$
 & \cellcolor[gray]{.8}$1.04\textsc{e}{-}04$
 & \cellcolor[gray]{.8}$8.87\textsc{e}{-}05$
 \\
\midrule
\multirow{2}{*}{DRM} & BC & $4.66\textsc{e}{-}01$
 & $4.93\textsc{e}{-}01$
 & $3.47\textsc{e}{-}01$
 & $2.30\textsc{e}{-}01$
 & $1.70\textsc{e}{-}01$
 \\
& FB & $1.01\textsc{e}{-}04$
 & $2.11\textsc{e}{-}04$
 & $2.77\textsc{e}{-}04$
 & $3.78\textsc{e}{-}04$
 & $3.10\textsc{e}{-}04$
 \\
\midrule
\multirow{2}{*}{WAN} & BC & $1.07\textsc{e}{-}01$
 & $1.93\textsc{e}{-}02$
 & $3.42\textsc{e}{-}02$
 & $3.86\textsc{e}{-}02$
 & $1.38\textsc{e}{-}01$
 \\
& FB & $1.12\textsc{e}{-}03$
 & $4.88\textsc{e}{-}04$
 & $5.62\textsc{e}{-}04$
 & $6.42\textsc{e}{-}04$
 & $8.01\textsc{e}{-}04$
 \\
\bottomrule
\end{tabular}
\end{table}

\subsection{Schrödinger Equation Baselines}
\subsubsection{One-Dimensional Infinite Potential Well}
\begin{figure}[htbp]  
    \centering
    \begin{subfigure}[t]{0.47\textwidth}  
        \centering
        \includegraphics[width=\textwidth,height=5cm,keepaspectratio]{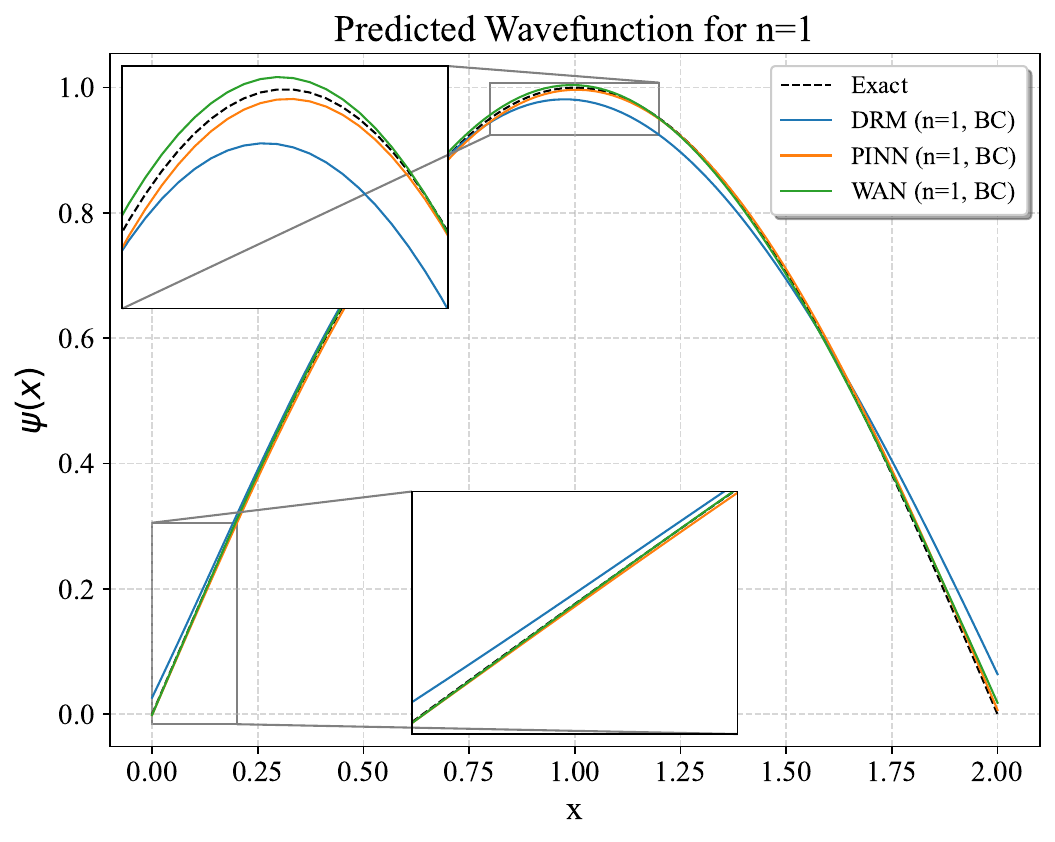}
        \caption{Predicted wavefunction for $n=1$.}
        \label{fig:wavefunction_n1}
    \end{subfigure}
    \hfill  
    \begin{subfigure}[t]{0.5\textwidth}
        \centering
        \includegraphics[width=\textwidth,height=5cm,keepaspectratio]{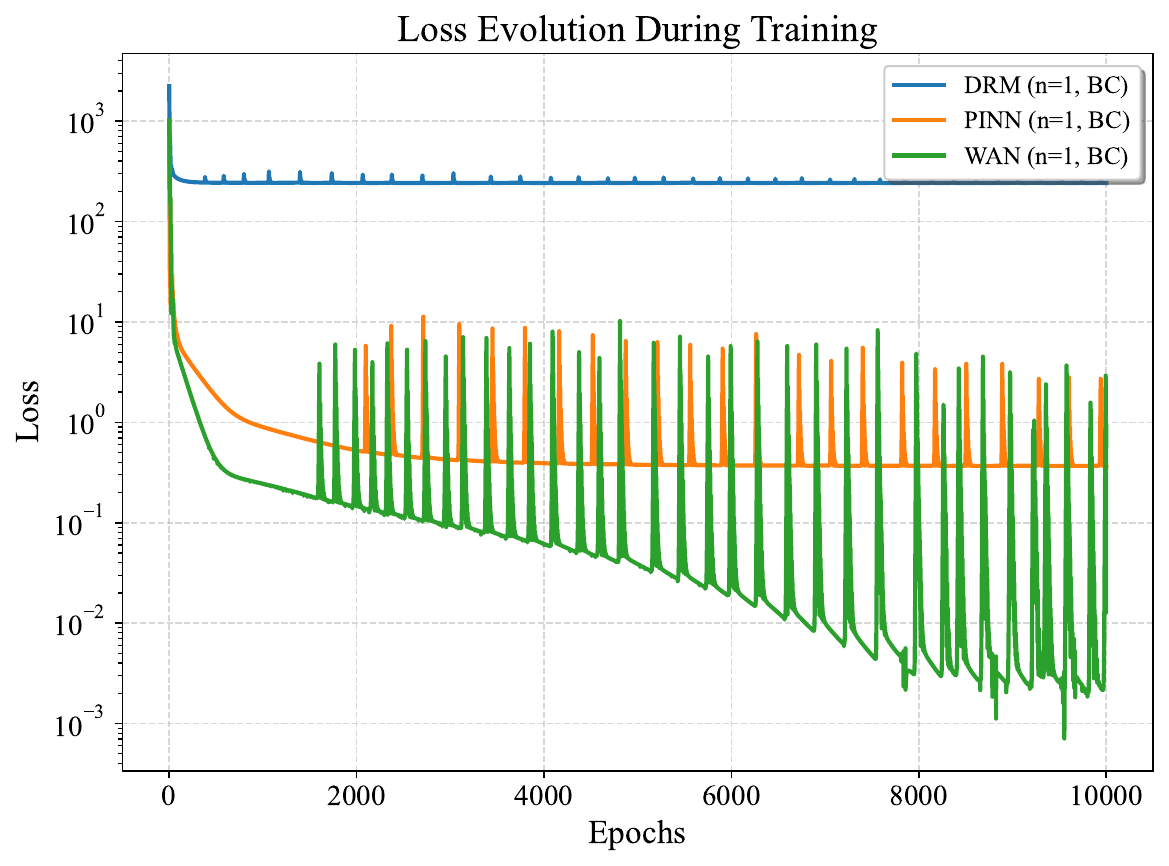}
        \caption{Loss evolution during training.}
        \label{fig:loss}
    \end{subfigure}
    \caption{Comparison of wavefunction predictions and loss for $n=1$.}
    \label{fig:combined_results}
\end{figure}

To benchmark our approach against established baselines, we compare it on the one-dimensional infinite potential well, examining states from the ground state ($n=1$) to the fourth excited state ($n=5$). We use `-' to denote methods that fail to converge, defined as cases where the $L_2$ error exceeds $1 \times 10^{-3}$, while run times are recorded for achieving the best $L_2$ error. These comparisons are detailed in Table~\ref{Baseline Comparison for Infinite Potential Well}. In Figure~\ref{fig:combined_results}, we illustrate the predicted wavefunction (Figure~\ref{fig:wavefunction_n1}) and loss evolution (Figure~\ref{fig:loss}) for each method.
\begin{table}[ht]
  \centering
  \caption{Baseline Comparison for Infinite Potential Well}
  \label{Baseline Comparison for Infinite Potential Well}
  \begin{tabular}{c *{6}{c}}
    \toprule
    & \multicolumn{2}{c}{PINN} 
    & \multicolumn{2}{c}{DRM} 
    & \multicolumn{2}{c}{WAN} \\
    \cmidrule(lr){2-3} \cmidrule(lr){4-5} \cmidrule(lr){6-7}
    $n$ & $L_2$ Error & Run Time & $L_2$ Error & Run Time & $L_2$ Error & Run Time \\
    \midrule
   $1$  & $6.22\textsc{e}{-}05$ & 21.6 & $4.62\textsc{e}{-}04$ & 19.2 & $2.16\textsc{e}{-}05$
 & 112.8 
 \\
   $2$ & $7.79\textsc{e}{-}06$
 & 105.3 
 & - & - & $1.03\textsc{e}{-}05$
 & 343.5 
 \\
    $3$  & $5.90\textsc{e}{-}07$
 & 19.0 
 & - & - & $4.72\textsc{e}{-}06$
 & 350.3 
 \\
    $4$  & $5.18\textsc{e}{-}06$
 & 88.0 
 & - & - & $1.33\textsc{e}{-}04$
 & 204.3 
 \\
    $5$  & $1.92\textsc{e}{-}07$
 & 55.4 
 & - & - & - & - \\
    \bottomrule
  \end{tabular}
\end{table}

In Figure~\ref{fig:combined_results}, we show the predicted wavefunction (Figure~\ref{fig:wavefunction_n1}) and loss evolution (Figure~\ref{fig:loss}) for different methods.

\subsubsection{One-Dimensional Quantum Harmonic Oscillator}
\begin{figure}[htbp]  
    \centering
    \begin{subfigure}[t]{0.47\textwidth}  
        \centering
        \includegraphics[width=\textwidth,height=5cm,keepaspectratio]{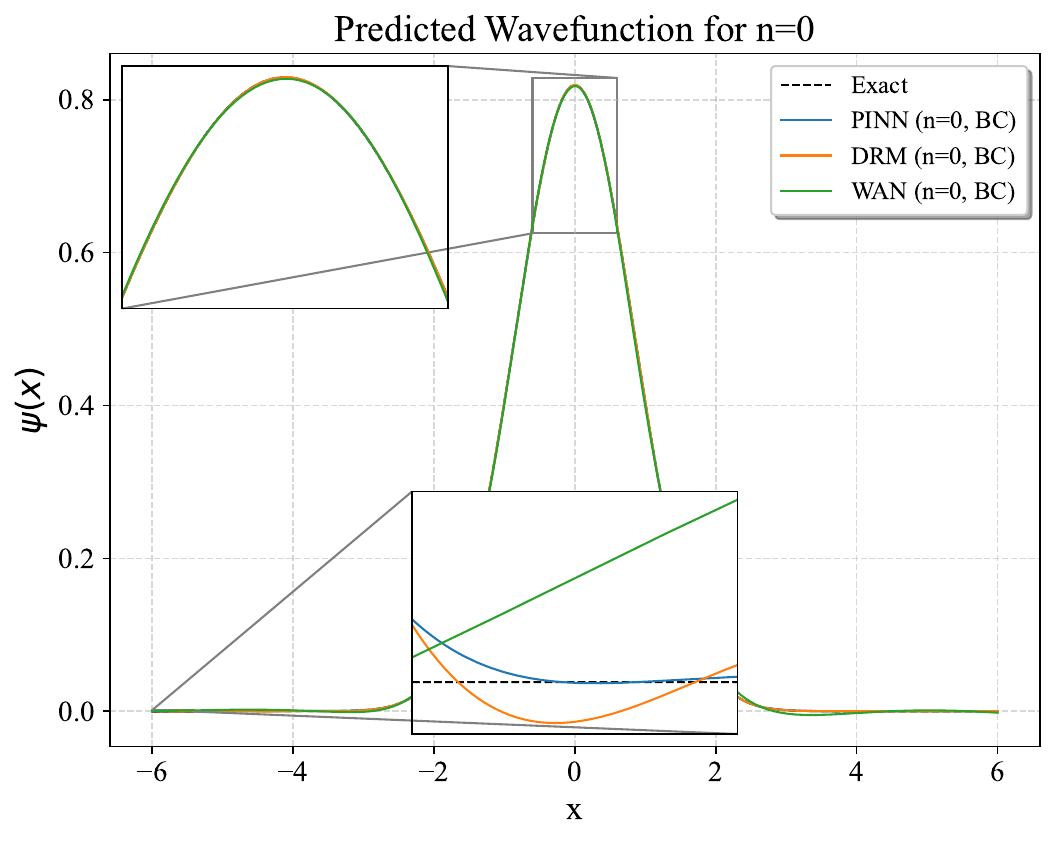}
        \caption{Predicted wavefunction for $n=1$.}
        \label{fig:wavefunction_n0_QHO}
    \end{subfigure}
    \hfill  
    \begin{subfigure}[t]{0.5\textwidth}
        \centering
        \includegraphics[width=\textwidth,height=5cm,keepaspectratio]{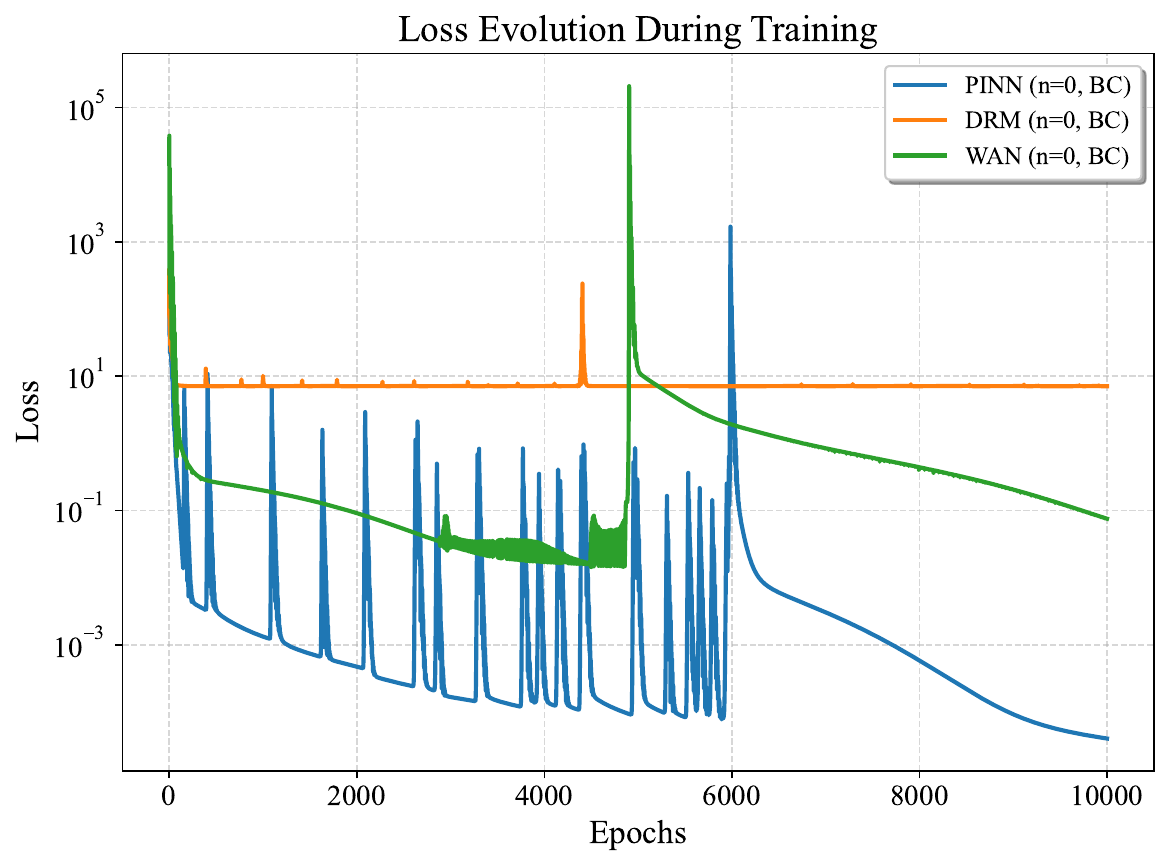}
        \caption{Loss evolution during training.}
        \label{fig:loss_QHO}
    \end{subfigure}
    \caption{Comparison of wavefunction predictions and loss for $n=0$.}
    \label{fig:combined_results_QHO}
\end{figure}

Similarly, we assess the baseline methods on the one-dimensional quantum harmonic oscillator, spanning states from the ground state ($n=0$) to the fourth excited state ($n=4$). As before, `-' indicates failure when the $L_2$ error is larger than $1 \times 10^{-3}$, with run times noted for the optimal $L_2$ error. The results are shown in Table~\ref{Baseline Comparison for Quantum Harmonic Oscillation}. Figure~\ref{fig:combined_results_QHO} shows the predicted wavefunction (Figure~\ref{fig:wavefunction_n0_QHO}) and loss evolution (Figure~\ref{fig:loss_QHO}) for the various methods.
\begin{table}[ht]
  \centering
  \caption{Baseline Comparison for Quantum Harmonic Oscillation}
  \label{Baseline Comparison for Quantum Harmonic Oscillation}
  \begin{tabular}{c *{6}{c}}
    \toprule
    & \multicolumn{2}{c}{PINN} 
    & \multicolumn{2}{c}{DRM} 
    & \multicolumn{2}{c}{WAN} \\
    \cmidrule(lr){2-3} \cmidrule(lr){4-5} \cmidrule(lr){6-7}
    $n$ & $L_2$ Error & Run Time & $L_2$ Error & Run Time & $L_2$ Error & Run Time \\
    \midrule
    $0$  & $1.30\textsc{e}{-}08$
 & 162.6 
 & $3.60\textsc{e}{-}08$
 & 67.1 

 & $3.84\textsc{e}{-}05$

 & 418.0 
 
 \\   $1$  & $2.15\textsc{e}{-}07$

 & 175.4 
 
 & - & - & $2.33\textsc{e}{-}04$

 & 138.2 

 \\
    $2$  & $1.12\textsc{e}{-}07$

 & 178.2 
 
 & - & - & $2.34\textsc{e}{-}04$

 & 224.2 

 \\
    $3$  & $1.02\textsc{e}{-}07$

 & 198.0 

 & - & - & -

 & -

 \\
    $4$  & $1.96\textsc{e}{-}07$

 & 179.3

 & - & - & - & - \\
    \bottomrule
  \end{tabular}
\end{table}

In Figure~\ref{fig:combined_results_QHO}, we show the predicted wavefunction\footnote{Further plots can be found in the Appendix.} (Figure~\ref{fig:wavefunction_n0_QHO}) and loss evolution (Figure~\ref{fig:loss_QHO}) for different methods. 

\section{Optimized Results and Improvements}

\subsection{Improvements on One-Dimensional Problems}
To enhance consistency in fitting higher states, we incorporate forced boundary and forced node conditions. This enables the neural network to better identify the specific state being fitted.
Table~\ref{Results for One Dimensional Infinite Potential Well} presents the results for the one-dimensional infinite potential well under forced boundary (FB), orthogonality (OG), and forced node (FN) conditions. The model was trained sequentially from the ground state ($n=1$) to the 5th excited state ($n=5$), with the best $L_2$ error recorded over \num{50000} epochs. Any $L_2$ errors exceeding $1 \times 10^{-3}$ are denoted by `-', indicating failure to converge on the solution. Shaded gray cells highlight the best performance in each case. Note that ground state results are identical across methods, as their initial conditions remain unaffected; only higher-order states are influenced by the OG and FN conditions.

\begin{table}[ht]
\centering
\caption{Results for One-Dimensional Infinite Potential Well}
\label{Results for One Dimensional Infinite Potential Well}
\begin{tabular}{l l  c c c c c }
\toprule
& & $n=1$ & $n=2$ & $n=3$ & $n=4$ & $n=5$ \\
\midrule
\multirow{2}{*}{PINN} & FB & $2.65\textsc{e}{-}06$ & $9.34\textsc{e}{-}06$ & $1.05\textsc{e}{-}06$ & $1.47\textsc{e}{-}07$ & $4.36\textsc{e}{-}08$\\
& FN & $2.65\textsc{e}{-}06$ & $4.71\textsc{e}{-}07$ & $5.24\textsc{e}{-}07$ & \cellcolor[gray]{.8}$7.87\textsc{e}{-}08$ &  \cellcolor[gray]{.8}$1.78\textsc{e}{-}08$\\
\midrule
\multirow{3}{*}{DRM} & FB & $1.84\textsc{e}{-}07$ & - & - & - & - \\
& OG & $1.84\textsc{e}{-}07$ & $3.92\textsc{e}{-}06$ & $4.51\textsc{e}{-}06$ & $5.72\textsc{e}{-}06$ & $6.88\textsc{e}{-}06$\\
& FN & $1.84\textsc{e}{-}07$ & $4.63\textsc{e}{-}07$ & $2.66\textsc{e}{-}06$ & $8.68\textsc{e}{-}05$ & $9.94\textsc{e}{-}05$\\
\midrule
\multirow{3}{*}{WAN} & FB & $8.79\textsc{e}{-}09$ & $1.67\textsc{e}{-}06$ & $2.11\textsc{e}{-}05$ & $7.36\textsc{e}{-}05$ & - \\
& OG & $8.79\textsc{e}{-}09$ & $1.30\textsc{e}{-}05$ & $2.38\textsc{e}{-}05$ & $1.29\textsc{e}{-}05$ & $3.75\textsc{e}{-}06$\\
& FN & \cellcolor[gray]{.8}$4.80\textsc{e}{-}09$ & \cellcolor[gray]{.8}$6.88\textsc{e}{-}09$ &  \cellcolor[gray]{.8}$8.50\textsc{e}{-}09$ & $2.56\textsc{e}{-}07$ & $3.14\textsc{e}{-}05$\\
\bottomrule
\end{tabular}
\end{table}

Using the same setup, Table~\ref{Results for One Dimensional Quantum Harmonic Oscillator} presents the optimized results for the one-dimensional quantum harmonic oscillator over \num{50000} epochs of training, starting from the ground state ($n=0$) to the fourth excited state. 
\begin{table}[ht]
\centering
\caption{Results for One-Dimensional Quantum Harmonic Oscillator}
\label{Results for One Dimensional Quantum Harmonic Oscillator}
\begin{tabular}{l l c c c c c}
\toprule
& & $n=0$ & $n=1$ & $n=2$ & $n=3$ & $n=4$ \\
\midrule
\multirow{2}{*}{PINN} & FB & $3.72\textsc{e}{-}08$
 & $1.19\textsc{e}{-}07$
 &  \cellcolor[gray]{.8}$1.81\textsc{e}{-}08$
 &  \cellcolor[gray]{.8}$1.42\textsc{e}{-}08$
 &  \cellcolor[gray]{.8}$1.73\textsc{e}{-}08$
 \\
& FN & $1.19\textsc{e}{-}07$ & $8.83\textsc{e}{-}07$
 & $6.06\textsc{e}{-}05$
 & $6.88\textsc{e}{-}04$
 &  - \\
\midrule
\multirow{3}{*}{DRM} & FB &  \cellcolor[gray]{.8}$9.99\textsc{e}{-}09$
 & - & - & - & - \\
& OG & $9.99\textsc{e}{-}09$ &  \cellcolor[gray]{.8}$7.38\textsc{e}{-}08$
 & $8.97\textsc{e}{-}08$
 & $8.15\textsc{e}{-}08$
 & $8.11\textsc{e}{-}08$
 \\
& FN & $9.99\textsc{e}{-}09$ & $2.18\textsc{e}{-}07$
 & $1.12\textsc{e}{-}05$
 & $1.94\textsc{e}{-}05$
 & - \\
\midrule
\multirow{3}{*}{WAN} & FB & $1.93\textsc{e}{-}05$
 & $8.59\textsc{e}{-}05$
 & $6.20\textsc{e}{-}04$
 & $7.61\textsc{e}{-}04$
 & - \\
& OG & $1.93\textsc{e}{-}05$
 & $5.38\textsc{e}{-}05$
 & $9.37\textsc{e}{-}05$
 & $5.89\textsc{e}{-}05$
 & $2.39\textsc{e}{-}04$
 \\
& FN & $1.93\textsc{e}{-}05$
 & $8.36\textsc{e}{-}04$ & - & - & - \\
\bottomrule
\end{tabular}
\end{table}

\subsection{Extensions to High-Dimensionality Problems}
Building on the optimization strategies applied in the one-dimensional case, we extend our approach to two-dimensional problems. Starting with the forced boundary condition, Table~\ref{Results for Two Dimensional Infinite Potential Well} shows the results for the two-dimensional infinite potential well. We investigate states ranging from the ground state (1,1) to (3,3), including mixed modes such as (2,1) and (3,1). Similarly, Table~\ref{Results for Two Dimensional Quantum Harmonic Oscillator} displays the corresponding results for the two-dimensional quantum harmonic oscillator, starting from the ground state (0,0).

\begin{table}[ht]
  \centering
  \caption{Results for Two-Dimensional Infinite Potential Well}
  \label{Results for Two Dimensional Infinite Potential Well}
  \begin{tabular}{c *{6}{c}}
    \toprule
    & \multicolumn{2}{c}{PINN} 
    & \multicolumn{2}{c}{DRM} 
    & \multicolumn{2}{c}{WAN} \\
    \cmidrule(lr){2-3} \cmidrule(lr){4-5} \cmidrule(lr){6-7}
    $n$ & FB & FN & OG & FN & OG & FN \\
    \midrule
    $(1,1)$  & $6.56\textsc{e}{-}10$
 & $6.56\textsc{e}{-}10$ & $7.99\textsc{e}{-}06$
 & $7.99\textsc{e}{-}06$ & $2.16\textsc{e}{-}05$
 & $6.61\textsc{e}{-}06$
 
 \\
    $(2,1)$  & $1.22\textsc{e}{-}08$

 & $2.83\textsc{e}{-}09$

 & $1.02\textsc{e}{-}05$
 & $4.85\textsc{e}{-}06$
 & $1.03\textsc{e}{-}05$
 & $4.85\textsc{e}{-}06$

 \\
    $(2,2)$  & $1.10\textsc{e}{-}07$

 & $7.23\textsc{e}{-}11$

 & - & $3.99\textsc{e}{-}06$
 & $4.72\textsc{e}{-}06$
 & $2.57\textsc{e}{-}05$
 
 \\
    $(3,1)$  & $2.04\textsc{e}{-}07$

 & $2.31\textsc{e}{-}09$

 & - & $6.88\textsc{e}{-}06$
 & $1.33\textsc{e}{-}04$
 & $1.24\textsc{e}{-}05$

 \\
    $(3,2)$  & $1.83\textsc{e}{-}06$

 & $4.92\textsc{e}{-}10$
 
 & - & $2.82\textsc{e}{-}05$
 & - & $9.35\textsc{e}{-}06$
 \\
    $(3,3)$  & $6.61\textsc{e}{-}05$

 & $1.39\textsc{e}{-}09$

 & - & $1.93\textsc{e}{-}05$
 & - & $1.31\textsc{e}{-}04$
 \\
    \bottomrule
  \end{tabular}
\end{table}

\begin{table}[ht]
  \centering
  \caption{Results for Two-Dimensional Quantum Harmonic Oscillator}
  \label{Results for Two Dimensional Quantum Harmonic Oscillator}
  \begin{tabular}{c *{6}{c}}
    \toprule
    & \multicolumn{2}{c}{PINN} 
    & \multicolumn{2}{c}{DRM} 
    & \multicolumn{2}{c}{WAN} \\
    \cmidrule(lr){2-3} \cmidrule(lr){4-5} \cmidrule(lr){6-7}
    $n$ & FB & FN & OG & FN & OG & FN \\
    \midrule
    $(0,0)$  & $6.71\textsc{e}{-}07$

 & $6.71\textsc{e}{-}07$
 & $7.99\textsc{e}{-}06$
 & $7.99\textsc{e}{-}06$ & $2.16\textsc{e}{-}05$
 & $6.61\textsc{e}{-}06$
 
 \\
    $(1,0)$  & $2.38\textsc{e}{-}06$

 & $3.12\textsc{e}{-}04$

 & $1.02\textsc{e}{-}05$
 & $4.85\textsc{e}{-}06$
 & $1.03\textsc{e}{-}05$
 & $4.85\textsc{e}{-}06$

 \\
    $(1,1)$  & $6.49\textsc{e}{-}05$

 & $2.69\textsc{e}{-}03$

 & - & $3.99\textsc{e}{-}06$
 & $4.72\textsc{e}{-}06$
 & $2.57\textsc{e}{-}05$
 
 \\
    $(2,0)$  & $2.09\textsc{e}{-}05$

 & $1.22\textsc{e}{-}03$

 & - & $6.88\textsc{e}{-}06$
 & $1.33\textsc{e}{-}04$
 & $1.24\textsc{e}{-}05$

 \\
    $(2,1)$  & $1.34\textsc{e}{-}04$

 & -

 & - & $2.82\textsc{e}{-}05$
 & - & $9.35\textsc{e}{-}06$
 \\
    $(2,2)$  & $9.51\textsc{e}{-}05$

 & -

 & - & $1.93\textsc{e}{-}05$
 & - & $1.31\textsc{e}{-}04$
 \\
    \bottomrule
  \end{tabular}
\end{table}

\section{Sensitivity and Ablations}
To demonstrate the data-free nature of our approach, we evaluate its performance in solving the time-independent Schrödinger equation without any training data points for both one-dimensional and two-dimensional systems, including the infinite potential well and quantum harmonic oscillator. The results encompass various eigenstates, highlighting the accuracy and convergence achieved solely through physics-informed constraints. These findings are summarized in Table~\ref{No-Data Training Results for 1D and 2D Schrödinger Systems}.

\begin{table}[ht]
  \centering
  \caption{No-Data Training Results for 1D and 2D Schrödinger Systems}
  \label{No-Data Training Results for 1D and 2D Schrödinger Systems}
  \begin{tabular}{c *{4}{c}}
    \toprule
    & \multicolumn{2}{c}{\text{IPW}} 
    & \multicolumn{2}{c}{\text{QHO}}  \\
    \cmidrule(lr){2-3} \cmidrule(lr){4-5} 
    $\text{method}$ & 1D & 2D & 1D & 2D \\
    \midrule
   PINN  & $4.03\textsc{e}{-}04$ & $2.45\textsc{e}{-}01$ & $3.53\textsc{e}{-}08$ & $2.20\textsc{e}{-}07$
 \\
  DRM  & $1.46\textsc{e}{-}07$
 & $2.55\textsc{e}{-}06$
 & $1.14\textsc{e}{-}08$ & $1.04\textsc{e}{-}07$
 \\
  WAN  & $1.68\textsc{e}{-}06$
 & $1.47\textsc{e}{-}06$
 & $1.93\textsc{e}{-}05$ & $1.04\textsc{e}{-}04$
 \\
    \bottomrule
  \end{tabular}
\end{table}

Furthermore, we conduct an ablation study to assess the sensitivity of our neural network architecture by varying the number of hidden layers from 1 to 4 and the number of neurons per layer among 10, 50, and 100. This analysis is performed on the one-dimensional infinite potential well at the fourth excited state ($n=5$), selected due to its increased complexity compared to the ground state, which allows for a more rigorous evaluation of architectural impacts on optimization and error metrics. The outcomes, including $L_2$ errors, are presented in Table~\ref{Ablation Study on Network Architecture for 1D Infinite Potential Well (n=5)}.
\begin{table}[ht]
\centering
\caption{Ablation Study on Network Architecture for 1D Infinite Potential Well ($n=5$)}
\label{Ablation Study on Network Architecture for 1D Infinite Potential Well (n=5)}
\begin{tabular}{l l c c c c }
\toprule
& & $1$ & $2$ & $3$ & $4$  \\
\midrule
\multirow{3}{*}{PINN} & 10& $3.98\textsc{e}{-}02$
 & $2.22\textsc{e}{-}02$

 & $2.90\textsc{e}{-}02$

 & $3.83\textsc{e}{-}02$

 \\
& 50 & $2.02\textsc{e}{-}04$
 & $3.28\textsc{e}{-}08$

 & $1.78\textsc{e}{-}08$

 & $9.93\textsc{e}{-}09$

 \\
 & 100 & $2.02\textsc{e}{-}05$
 & $1.37\textsc{e}{-}08$

 & $9.42\textsc{e}{-}09$

 & $8.64\textsc{e}{-}09$

 \\
\midrule
\multirow{3}{*}{DRM} & 10 & $4.77\textsc{e}{-}02$

 & $1.68\textsc{e}{-}01$
 & $3.59\textsc{e}{-}02$
 & $4.52\textsc{e}{-}02$
 \\
& 50 & $1.33\textsc{e}{-}01$
 & $3.29\textsc{e}{-}06$

 & $4.96\textsc{e}{-}07$

 & $1.33\textsc{e}{-}07$

 \\
& 100 & $6.34\textsc{e}{-}04$
 & $7.56\textsc{e}{-}07$

 & $1.34\textsc{e}{-}07$

 & $1.13\textsc{e}{-}07$

 \\
\midrule
\multirow{3}{*}{WAN} & 10 & $1.47\textsc{e}{-}01$
 & $1.33\textsc{e}{-}02$
 & $5.74\textsc{e}{-}03$
 & $6.29\textsc{e}{-}04$
  \\
&50 & $9.71\textsc{e}{-}06$
 & $1.43\textsc{e}{-}01$
 & $7.23\textsc{e}{-}07$
 & $4.06\textsc{e}{-}04$
 
 \\
& 100 & $3.39\textsc{e}{-}05$
 & $4.06\textsc{e}{-}04$ & $2.27\textsc{e}{-}04$ & $2.47\textsc{e}{-}03$\\
\bottomrule
\end{tabular}
\end{table}

\section{Summary}

This chapter presents the empirical results from the experiments described in Chapter 3. Using a comprehensive evaluation protocol, we compare baseline and optimized implementations of PINN, DRM, and WAN on the Poisson equation and canonical Schrödinger problems (1D/2D infinite well, 1D/2D harmonic oscillator). Baselines expose common failure modes, such as trivial solutions, spectral bias, and elevated $L_2$ errors, while optimized variants incorporate targeted improvements (e.g., additional loss terms, hard boundary enforcement, and enhanced architectures) that markedly reduce these issues. Together, these results enable a fair, like-for-like comparison and clarify the strengths, limitations, and trade-offs of each method. A detailed analysis and discussion of implications is provided in Chapter 6.

\chapter{Schrödinger Equation with Kratzer-Hellmann Potential}
The interaction of intense laser fields with quantum systems gives rise to rich dynamical behaviour, often requiring nonperturbative treatment. A particularly useful framework for addressing atoms or electrons in strong oscillatory fields is the Kramers-Henneberger (KH) transformation, which moves into a non-inertial frame co-moving with the classical quiver motion induced by the laser. In this chapter, we formulate the time-dependent Schrödinger equation with the KH potential and develop a neural network-based solver to approximate its solution. This chapter comprises: (i) theoretical background and derivation of the KH potential, (ii) a presentation of numerical experiments and results, and (iii) a summary. Comparison and broader discussion with other PDEs and methods will be deferred to Chapter 6.

\section{Theoretical Background}
When a quantum particle (e.g., an electron bound in an atom or molecule) is subjected to a strong, oscillatory electromagnetic field, such as intense laser light, the interaction can drive a large-amplitude quivering motion. In the laboratory frame, this manifests as a time-dependent forcing term that makes the Schrödinger equation rapidly oscillatory and, in strong-field regimes, nonperturbative. A powerful way to tame and reinterpret these dynamics is to move into a non-inertial frame that co-moves with the classical oscillatory displacement induced by the laser. This is the essence of the Kramers-Henneberger (KH) transformation: instead of treating the laser as an external time-dependent field acting on a stationary binding potential, one views the binding potential as oscillating in time while the electron is `stationary' in the moving frame. In many circumstances, this recasting leads to clearer physical intuition and simplifications such as effective potentials and stabilization phenomena. 

\subsection{Classical Quivering and Frame Change}
Consider a free electron driven by a linearly polarized laser field in the dipole approximation. The classical equation of motion (ignoring magnetic and relativistic effects) is
\begin{equation}
m \ddot{\boldsymbol{\alpha}}(t)=-\mathbf{F}(t)
\end{equation}
where $\mathbf{F}(t)$ is the electric field of the laser. Integrating twice gives the quivering displacement
\begin{equation}
\boldsymbol{\alpha}(t)=\frac{1}{m \omega^2} F_0 \sin (\omega t) \hat{\mathbf{e}}.
\end{equation}

This is the classical oscillatory shift that a free electron undergoes under laser excitation.
The KH transformation is a time-dependent unitary change of variables that shifts coordinates by this classical displacement. We define a new wavefunction in the KH frame:
\begin{equation}
\Psi(\mathbf{r}, t)=\exp \left[-\frac{i}{\hbar} m \dot{\boldsymbol{\alpha}}(t) \cdot \mathbf{r}\right] \Phi(\mathbf{r}+\boldsymbol{\alpha}(t), t).
\end{equation}
In physical terms, this amounts to going into the accelerating frame following the quivering and applying the appropriate phase to maintain unitarity. Substituting this into the original time-dependent Schrödinger equation with a binding potential $V(\mathbf{r})$ and the laser field in the length gauge transforms the direct laser coupling into a time-dependent shifted potential:
\begin{equation}
i \hbar \frac{\partial}{\partial t} \Phi(\mathbf{r}, t)=\left[-\frac{\hbar^2}{2 m} \nabla^2+V(\mathbf{r}+\boldsymbol{\alpha}(t))\right] \Phi(\mathbf{r}, t)
\end{equation}
That is, in the KH frame, the electron evolves under a potential that oscillates in time according to the classical quiver trajectory, known as the Kramers-Henneberger potential:
\begin{equation}
V_{\mathrm{KH}}(\mathbf{r}, t):=V(\mathbf{r}+\boldsymbol{\alpha}(t)) .
\end{equation}

The laser field no longer appears explicitly as a driving term; its effect is encoded in the oscillatory shift of the binding potential.

\subsection{Cycle Averaging and Effective Potential}
If the driving laser is periodic with angular frequency $\omega$, i.e., $\boldsymbol{\alpha}(t+T)=\boldsymbol{\alpha}(t)$ with $T=2 \pi / \omega$, and the frequency is large compared to the intrinsic energy scales of the bound system, then the rapid oscillations of the KH potential can be smoothed out by averaging over one period. We define the cycle-averaged potential as
\begin{equation}
\bar{V}_{\mathrm{KH}}(\mathbf{r}):=\frac{1}{T} \int_0^T V(\mathbf{r}+\boldsymbol{\alpha}(t)) \ d t.
\end{equation}
Under this approximation, the time-dependent KH Schrödinger equation
\begin{equation}
i \hbar \frac{\partial}{\partial t} \Phi(\mathbf{r}, t)=\left[-\frac{\hbar^2}{2 m} \nabla^2+V(\mathbf{r}+\boldsymbol{\alpha}(t))\right] \Phi(\mathbf{r}, t)
\end{equation}
is approximated by the time-independent eigenvalue equation
\begin{equation}
\left[-\frac{\hbar^2}{2 m} \nabla^2+\bar{V}_{\mathrm{KH}}(\mathbf{r})\right] \psi(\mathbf{r})=E \psi(\mathbf{r})
\end{equation}
where $E$ is interpreted as the leading-order quasi-energy of the dressed state. This approximation corresponds to keeping the zeroth-order term in a high-frequency (Floquet-Magnus or van Vleck) expansion of the effective Hamiltonian. Corrections to this are systematically computable as higher-order (in $1 / \omega$) terms that reintroduce residual time dependence.

\subsection{Problem Setup}
We consider the short-range potential~\cite{tasnim_aynul_quantum_2025}
\begin{equation}
V(x)=V_0\,\frac{\exp\!\left[-\sqrt{x^2+16}\right]}{\sqrt{x^2+6.27^2}},
\end{equation}
with $V_0=-24.856$ a.u.

\begin{figure}
    \centering
    \includegraphics[width=0.75\linewidth]{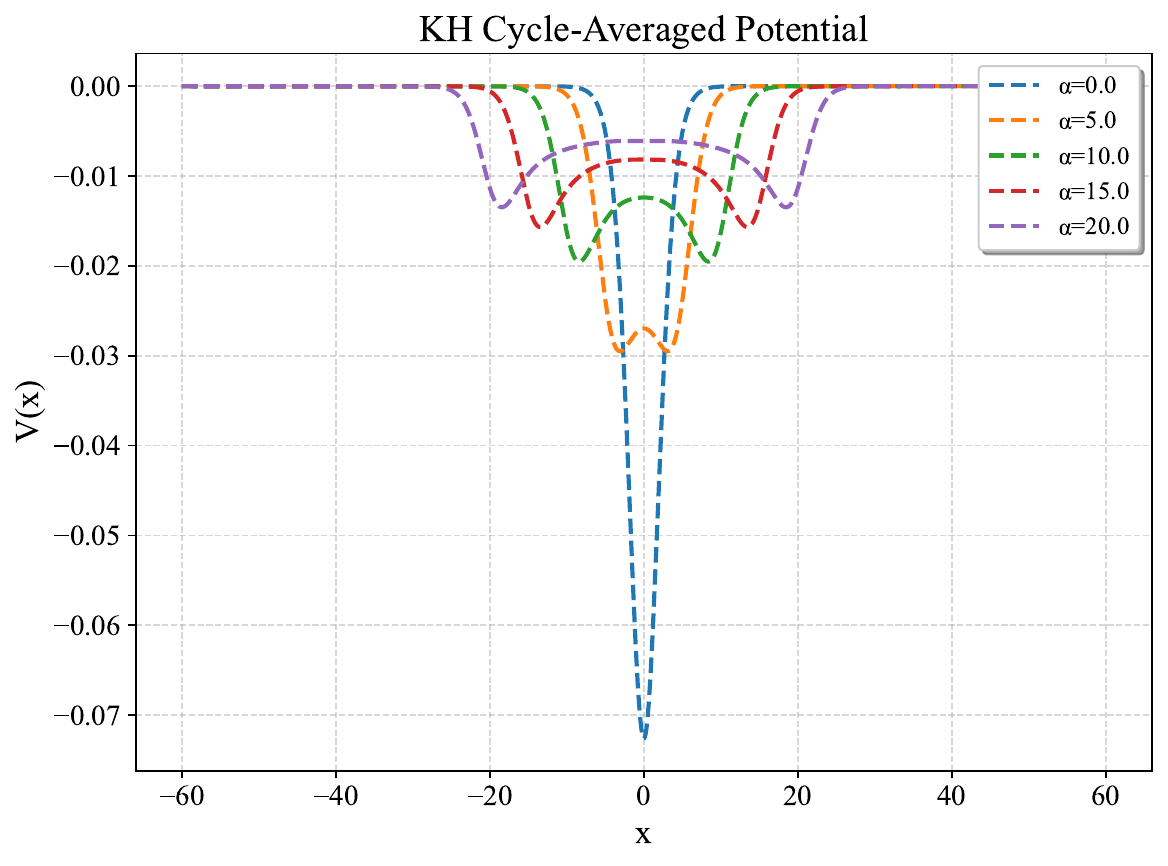}
    \caption{KH cycle-averaged potential.}
    \label{fig:kh_cycle_avg_potential}
\end{figure}

Averaging over one driving period yields the cycle-averaged KH potential shown in Fig.~\ref{fig:kh_cycle_avg_potential}. In our study, we solve the time-independent Schr\"odinger equation with $\alpha=10$ for the ground state through to the third excited states. We then perform a parameter sweep over $\alpha$.

For evaluation, we obtain reference (“ground-truth”) solutions using a finite difference method.

\section{Results}
We begin with the best-performing configuration from prior tests on other Schr\"odinger problems and train for \num{10000} epochs. Using $25\%$ and $50\%$ of the training data, we report results at $\alpha=10$ for the ground to third excited states.

\begin{table}[ht]
\centering
\caption{Results for the one-dimensional KH Schr\"odinger equation.}
\label{tab:kh_1d_results}
\begin{tabular}{l l c c c c c}
\toprule
& Data & $n=0$ & $n=1$ & $n=2$ & $n=3$ & \text{Run time}\\
\midrule
\multirow{2}{*}{PINN} & $25\%$ &  - & - & - & - & - \\
& $50\%$& $8.10\textsc{e}{-}09$ & $2.67\textsc{e}{-}09$ & $3.96\textsc{e}{-}09$ & $8.27\textsc{e}{-}09$ & $107$ \\
\midrule
\multirow{2}{*}{DRM} & $25\%$ & $1.47\textsc{e}{-}07$ & $1.86\textsc{e}{-}08$ & $1.37\textsc{e}{-}06$ & $3.56\textsc{e}{-}08$ & $101$ \\
& $ 50\% $& $1.74\textsc{e}{-}09$ & $2.29\textsc{e}{-}09$ & $4.51\textsc{e}{-}09$ & $9.04\textsc{e}{-}09$ & $102$ \\
\midrule
\multirow{2}{*}{WAN} & $25\%$ & - & - & - & - & - \\
& $50\%$ & $2.08\textsc{e}{-}08$ & $9.16\textsc{e}{-}09$ & $6.30\textsc{e}{-}09$ & $1.34\textsc{e}{-}08$ & $270$ \\
\bottomrule
\end{tabular}
\end{table}

Using $50\%$ of the training data, we next sweep $\alpha$ from $0$ to $20$ for the ground state only.

\begin{table}[ht]
\centering
\caption{Ground-state results across different values of $\alpha$.}
\label{tab:alpha_sweep_ground}
\setlength{\tabcolsep}{5pt}
\begin{tabular}{l c c c c c c}
\toprule
  & $\alpha=0$ & $\alpha=5$ &  $\alpha=10$ &  $\alpha=15$ & $\alpha=20$ & \text{Run time}\\
\midrule
PINN & $1.73\textsc{e}{-}08$ & $5.86\textsc{e}{-}09$ & $4.92\textsc{e}{-}09$ & $4.72\textsc{e}{-}09$ & $1.55\textsc{e}{-}08$ & $114$ \\
\midrule
DRM & $4.02\textsc{e}{-}09$ & $4.08\textsc{e}{-}09$ & $8.00\textsc{e}{-}09$ & $4.46\textsc{e}{-}09$ & $3.69\textsc{e}{-}09$ & $91$ \\
\midrule
WAN & $7.95\textsc{e}{-}08$ & $2.04\textsc{e}{-}08$ & $6.01\textsc{e}{-}09$ & $5.66\textsc{e}{-}09$ & $2.92\textsc{e}{-}09$ & $460$ \\
\bottomrule
\end{tabular}
\end{table}

\begin{figure}[htbp]
    \centering
    \begin{subfigure}[t]{0.48\textwidth}
        \centering
        \includegraphics[width=\textwidth,height=5cm,keepaspectratio]{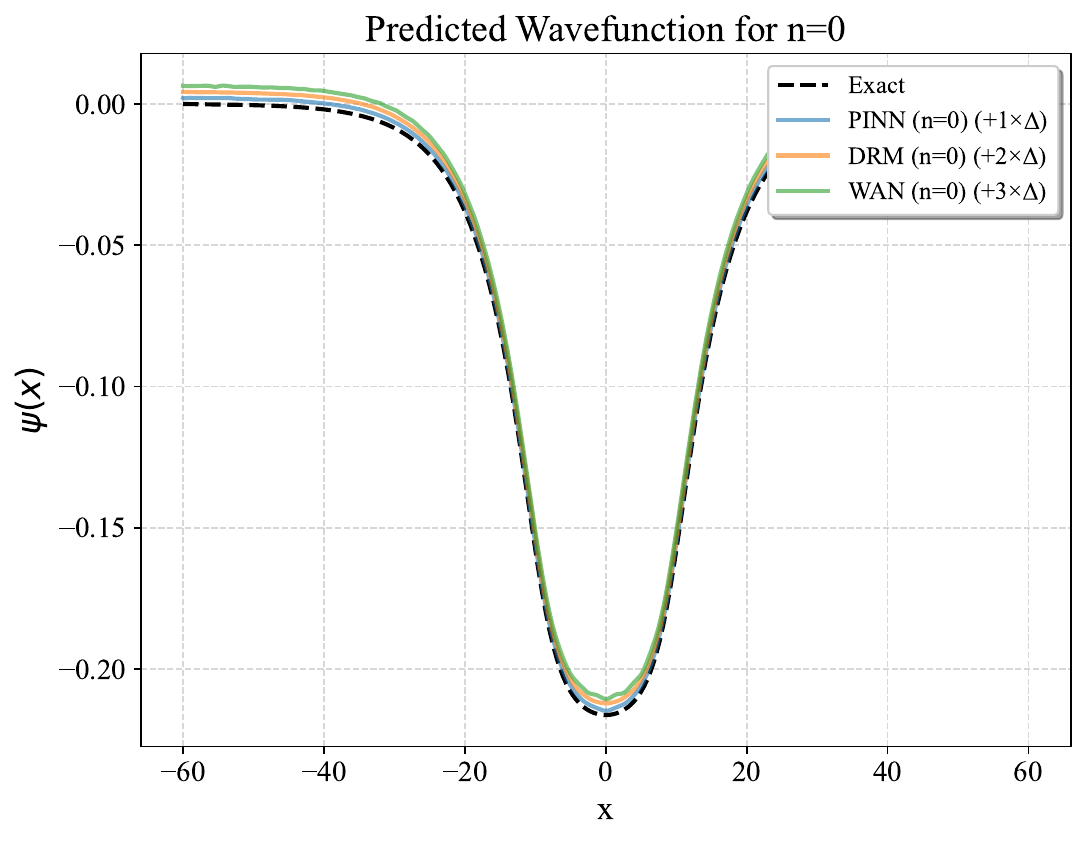}
        \label{fig:wavefunction_n0_KH}
    \end{subfigure}
    \hfill
    \begin{subfigure}[t]{0.48\textwidth}
        \centering
        \includegraphics[width=\textwidth,height=5cm,keepaspectratio]{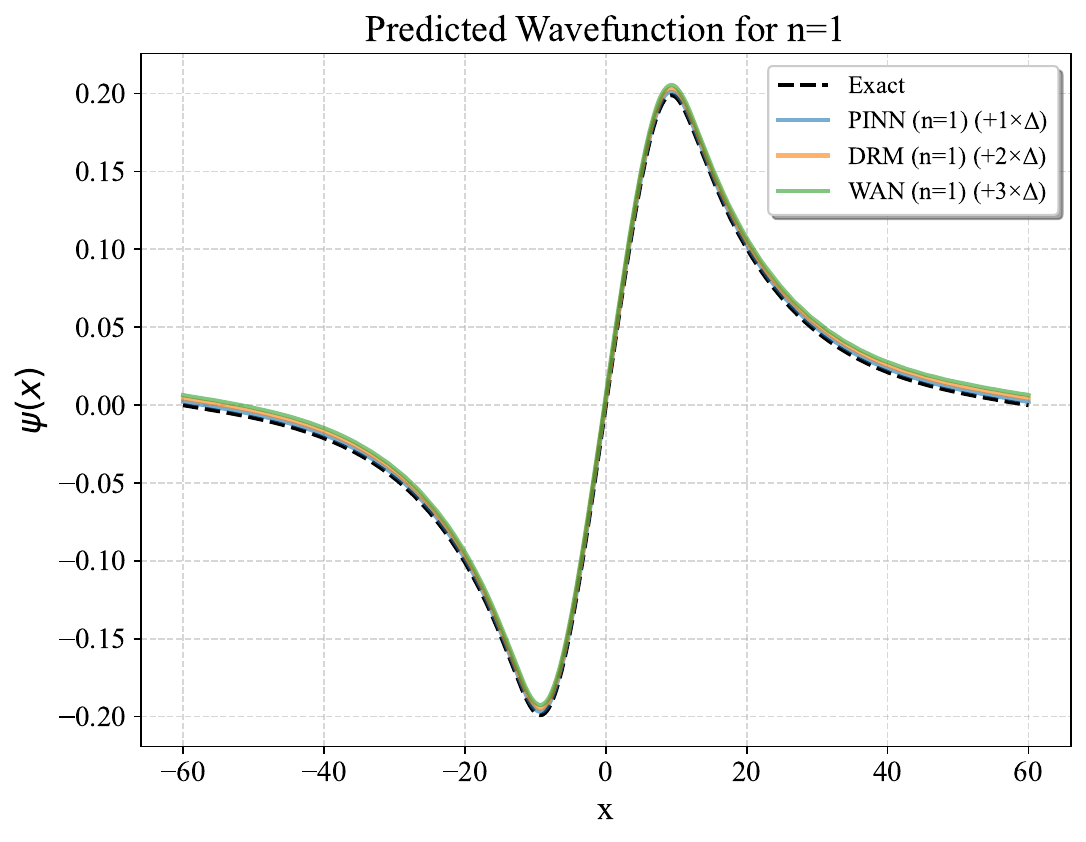}
        \label{fig:wavefunction_n1_KH}
    \end{subfigure}

    \vspace{0.1cm}

    \begin{subfigure}[t]{0.48\textwidth}
        \centering
        \includegraphics[width=\textwidth,height=5cm,keepaspectratio]{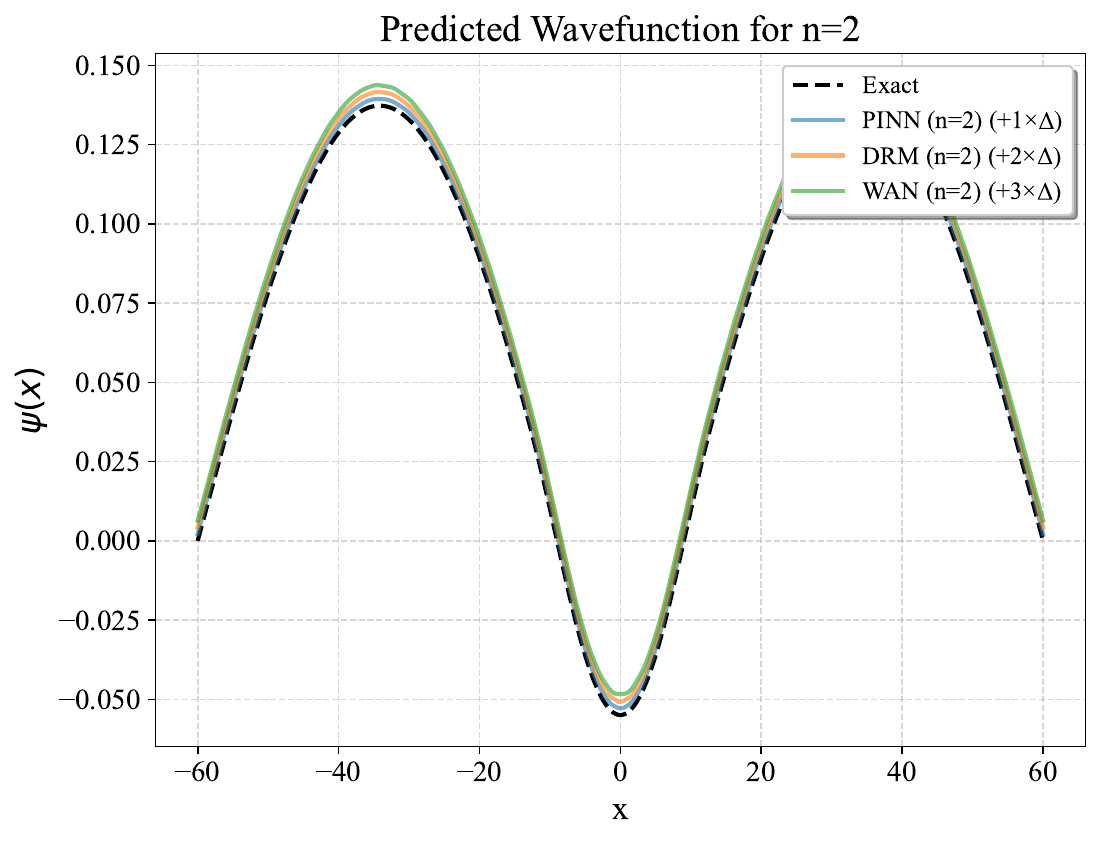}
        \label{fig:wavefunction_n2_KH}
    \end{subfigure}
    \hfill
    \begin{subfigure}[t]{0.48\textwidth}
        \centering
        \includegraphics[width=\textwidth,height=5cm,keepaspectratio]{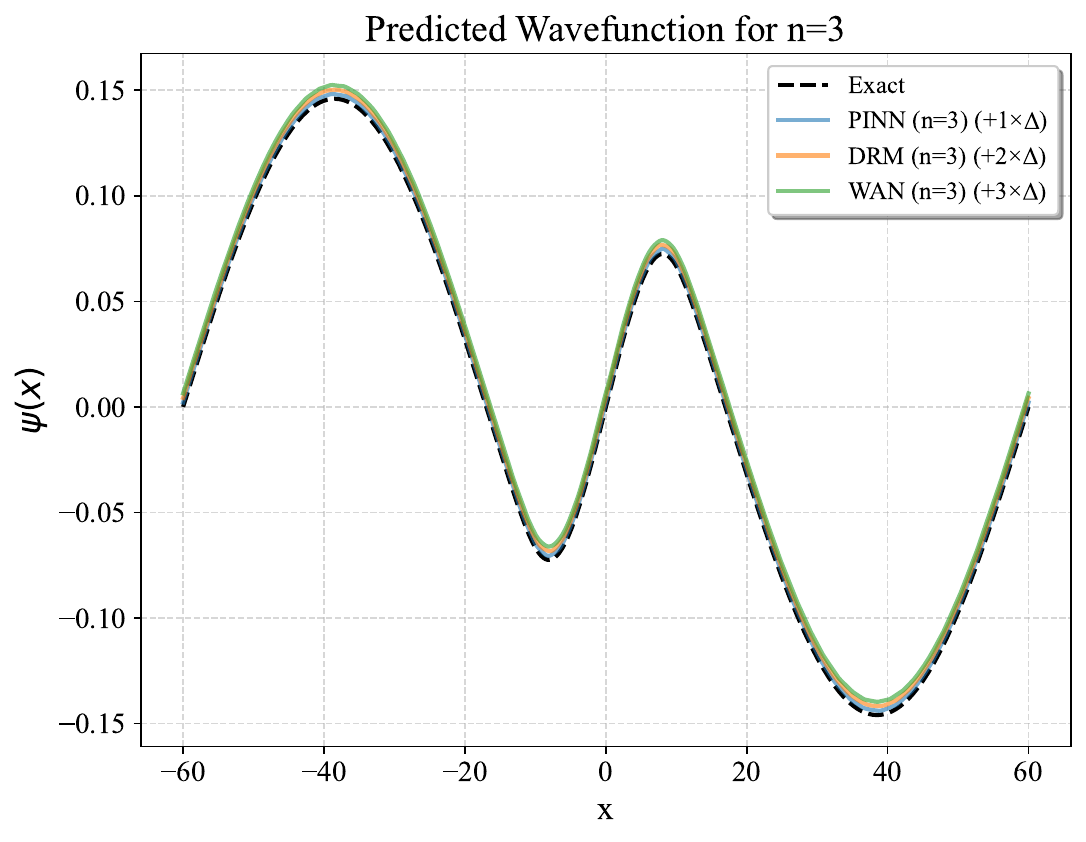}
        \label{fig:wavefunction_n3_KH}
    \end{subfigure}

    \caption{Predicted wavefunctions for $n=0$ to $n=3$ under the KH potential.}
    \label{fig:combined_results_KH}
\end{figure}

Figure~\ref{fig:combined_results_KH} shows the predicted wavefunctions for $n=0$ to $n=3$. Because all three methods achieve relatively small $L_2$ errors, small vertical offsets are applied to the curves for clearer visualisation.

\section{Summary}
In this chapter, we introduced an end-to-end neural network solver for the time-dependent Schr\"odinger equation formulated in the Kramers-Henneberger (KH) frame. Leveraging the KH transformation to move into a non-inertial frame co-moving with the laser-driven quivering, we derived the one-dimensional, cycle-averaged KH potential and cast the stationary problem into an eigenvalue-learning task suitable for neural approximation.

We instantiated and compared several neural PDE solvers, most notably physics-informed neural networks (PINNs), deep Ritz methods (DRM), and weak-adversarial approaches (WAN), under controlled training budgets.Solutions were validated against finite-difference reference data. At a representative coupling $\alpha=10$, the solver accurately recovered the ground through third excited states; we further examined robustness via a sweep over $\alpha\in\{0,5,10,15,20\}$ for the ground state. Across these studies, the predicted wavefunctions matched the reference solutions with consistently small $L_2$ errors while maintaining practical training times. A broader comparison with other PDE families and solver classes, including discussions of scalability and generalisation, will be provided in Chapter~6.

\chapter{Discussion}
This chapter synthesizes the results from Chapters~4 and~5 to assess three neural PDE solvers, physics-informed neural networks (PINNs), the deep Ritz method (DRM), and weak adversarial networks (WANs), on the benchmark problems defined in Chapters~3 and~5. Our objectives are to identify the strengths and limitations of each method across tasks, to explain failure modes and how targeted adjustments improve outcomes, and to provide practical guidance for choosing and configuring these methods for new problems. We begin with the Poisson equation and compare performance across spatial dimensions while examining training factors such as boundary-condition enforcement (soft penalties versus forced boundary conditions, FBC), sampling strategies, network architectures, and loss weighting. We then analyze the time-independent Schr\"odinger equation in one and two dimensions (infinite potential well and harmonic oscillator), interpreting the mechanisms behind the observed behaviours. Throughout, we cross-reference training strategies and distil recommendations applicable to similar PDE problems. Finally, we incorporate the time-dependent Kramers-Henneberger (KH) formulation as a demanding case study to probe robustness and physical fidelity.

\section{Key Findings}

Across the Poisson benchmarks, FBC consistently outperforms soft, weighted boundary penalties. Under FBC, PINN achieves the lowest \(L_2\) error at a fixed budget of \num{10000} epochs, whereas DRM has the shortest runtime. Because all three methods solve the multi-dimensional Poisson setups without marked degradation, there is no evidence of a dimensionality bottleneck under our settings: performance remains stable as the dimensionality increases up to five (see Table~\ref{Results for Multi-dimensional Poisson Equation}). Given its accuracy-speed balance, DRM is an attractive default when moderate accuracy and fast fits are required, while PINN is preferable when the priority is the lowest error within a fixed training budget.

For the Schr\"odinger benchmarks, with soft boundary conditions and only 25\% of domain data, all methods recover the ground state accurately in the infinite potential well (IPW) (Table~\ref{Baseline Comparison for Infinite Potential Well}). As we move to higher excited states, DRM collapses to the ground state, PINN continues to recover successive states reliably, and WAN succeeds up to the fourth excited state (\(n=5\)) before failing. Similar patterns appear in the quantum harmonic oscillator (QHO) baseline (Table~\ref{Baseline Comparison for Quantum Harmonic Oscillation}). Overall, PINN consistently attains the best accuracy across states, while DRM typically converges fastest but often only for the ground state under baseline settings. Representative runtimes illustrate this trade-off: for IPW, DRM requires \(19.2\)~s versus \(21.6\)~s for PINN; for the QHO, DRM requires \(67.1\)~s versus \(162.6\)~s for PINN.

Targeted improvements materially change this picture. Replacing soft boundary penalties with FBC improves all three methods solving the 1D IPW problem: the PINN \(L_2\) error drops from \(6.22\times 10^{-5}\) to \(2.65\times 10^{-6}\); DRM improves from \(4.62\times 10^{-4}\) to \(1.84\times 10^{-7}\); and WAN improves from \(2.16\times 10^{-5}\) to \(8.79\times 10^{-9}\) (Table~\ref{Results for One Dimensional Infinite Potential Well}). Adding orthogonality regularization (OG) stabilizes excited-state recovery: using the DRM, the per-state differences are small but overall accuracy is high, while in WAN the OG term yields strong \(L_2\) performance on higher states (Table~\ref{Results for One Dimensional Quantum Harmonic Oscillator}). Enforcing fixed/forced nodes (FN), i.e., known nodal locations, is particularly effective in the IPW problems (both 1D and 2D) where node spacing is regular; under FN, WAN achieves the best \(L_2\) error for \(n=1\)–\(3\), and PINN is best for \(n=4,5\) (Tables~\ref{Results for Two Dimensional Infinite Potential Well} and~\ref{Results for Two Dimensional Quantum Harmonic Oscillator}). FN is less effective in QHO due to non-uniform node spacing and it requires prior nodal information, which limits generality. In two-dimensional Schr\"odinger problems, degeneracies (e.g., \((1,2)\) versus \((2,1)\)) complicate OG; penalties can inadvertently mix degenerate states unless guided by initial data or symmetry cues. Our 2D experiments succeed at low states using data-driven disambiguation, but OG degrades at higher degeneracy levels.

\section{Cross-Method Comparison}

The outcomes from the various methods are summarised below for orientation; the measures themselves were given in previous chapters (Poisson in Table~\ref{Results for Multi-dimensional Poisson Equation}; IPW/QHO baselines in Tables~\ref{Baseline Comparison for Infinite Potential Well}–\ref{Baseline Comparison for Quantum Harmonic Oscillation}; improved settings in Tables~\ref{Results for One Dimensional Infinite Potential Well}–\ref{Results for Two Dimensional Quantum Harmonic Oscillator}).

\begin{table}[ht]
  \centering
  \caption{Cross-method summary by task and dimension. ``Accuracy'' reflects the lowest \(L_2\) among methods; ``Runtime'' the shortest Runtime.}
  \label{tab:cross-method}
  \begin{tabular}{lccccc}
    \toprule
    & \multicolumn{1}{c}{Poisson} & \multicolumn{2}{c}{IPW} & \multicolumn{2}{c}{QHO} \\
    \cmidrule(lr){2-2} \cmidrule(lr){3-4} \cmidrule(lr){5-6}
    Metric & 1D-5D & 1D & 2D & 1D & 2D \\
    \midrule
    Accuracy & {PINN} & {PINN} & {PINN} & {PINN} & {DRM} \\
    Runtime  & {DRM}  & {DRM}  & {DRM}  & {DRM}  & {DRM} \\
    \bottomrule
  \end{tabular}
\end{table}

In the Poisson family, adopting FBC removes sensitivity to the boundary-loss weight and improves stability across methods (Table~\ref{Results for Multi-dimensional Poisson Equation}). In Schr\"odinger problems, FBC yields modest gains for PINN (whose baselines already satisfy BCs) but large gains for DRM and WAN. However, FBC alone does not resolve high-state failures. FN is effective for the IPW, where nodes are regularly spaced, but unreliable for the QHO, where nodes are irregular. OG injects physical structure but is sensitive to degeneracy and to the accuracy of lower-state approximations. Treating the energy as a trainable parameter modestly improves PINN and DRM in QHO problems, while WAN appears hindered, likely due to generator-discriminator dynamics (compare the no-data results in Table~\ref{No-Data Training Results for 1D and 2D Schrödinger Systems} with the supervised baselines in Tables~\ref{Baseline Comparison for Infinite Potential Well}–\ref{Baseline Comparison for Quantum Harmonic Oscillation}).

\section{Sensitivity, Ablations, and Reproducibility}

\begin{figure}[htbp]
  \centering
  \includegraphics[width=\textwidth,height=5cm,keepaspectratio]{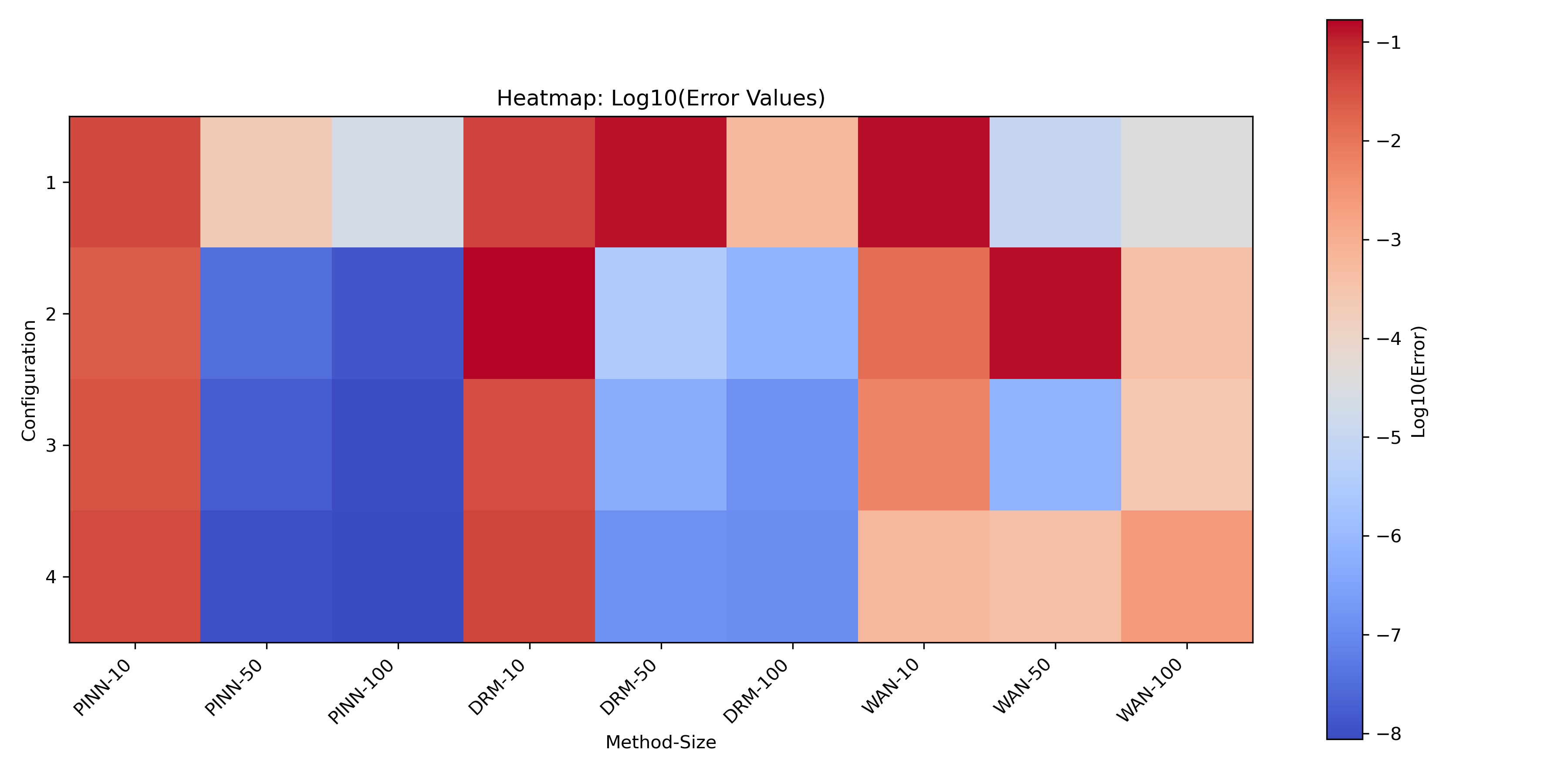}
  \caption{Error heatmap across architectures for \(n=1\), illustrating sensitivity to width and depth (lower is better).}
  \label{heatmap_errors}
\end{figure}

The experiments reveal clear sensitivities to loss weighting and supervision. Increasing the data-fit weight (partial supervision) improves baselines, whereas over-weighting the PDE residual degrades performance. In no-data settings focused on ground states, DRM outperforms others with \(L_2\) errors in the \(10^{-6}\)-\(10^{-8}\) range; PINN fails on the 2D IPW under these constraints while succeeding in 1D; and WAN matches DRM on the IPW but is weaker on QHO problems (Table~\ref{No-Data Training Results for 1D and 2D Schrödinger Systems}). Most of the achievable accuracy is reached within roughly \num{10000} epochs; extending to \num{50000} yields diminishing returns (for example, improvements from \(10^{-8}\) to \(10^{-9}\)). For practical runs, \num{5000} epochs often deliver \(10^{-4}\)–\(10^{-5}\) \(L_2\) error. Increasing the number of neurons per layer consistently improves PINN and DRM (single-network methods). For WAN, this trend holds only in shallow, one-layer settings; deeper settings require careful generator–discriminator width balance (our default used half the generator width for the discriminator). Increasing depth helps PINN and DRM modestly; for WAN, three layers perform best, and both too shallow and too deep configurations harm stability. These sensitivities are summarized in Table~\ref{Ablation Study on Network Architecture for 1D Infinite Potential Well (n=5)} and visualized in Fig.~\ref{heatmap_errors}.

The ablation studies further clarify causal contributions. Removing FBC and reverting to soft BCs increases boundary residuals and destabilizes training; dropping OG leads to mode mixing in excited states; and disabling FN especially harms IPW excited-state accuracy. Each configuration was repeated five times with fixed seeds, and variability was small to negligible. Seeds are fixed in the repository for deterministic sampling of collocation and boundary points. In our test problems, uniform sampling outperformed purely random sampling and produced more stable training, an effect visible in the consistency across repetitions in Table~\ref{No-Data Training Results for 1D and 2D Schrödinger Systems} and in the architecture-level robustness in Table~\ref{Ablation Study on Network Architecture for 1D Infinite Potential Well (n=5)}.

\section{KH Potential - Discussion}

Guided by the insights from solving the IPW and QHO problems, we next consider the KH frame for strong-field physics. The potential and wavefunction in this setting exhibit complex structure, and training is correspondingly less smooth. With OG and only 25\% of domain data, DRM alone converges reliably; increasing to 50\% domain coverage enables all three methods to perform well. PINN and DRM have comparable runtimes, whereas WAN is approximately twice as slow. When sweeping the KH parameter \(\alpha\), ground-state experiments show DRM achieving the best accuracy-speed trade-off, and the qualitative trends mirror the improvements observed under FBC and OG in the simpler benchmarks. 

\section{Practitioner-Oriented Guidelines}

Translating the comparison into practice, we recommend selecting the method by matching problem traits to solver strengths. Problems with simple spectra and regular nodal structure (e.g., IPW) benefit from DRM or WAN augmented by FN; problems with irregular nodal spacing or pronounced degeneracy (e.g., QHO, higher-dimensional Schrödinger) are better served by PINN with FBC and OG. When supervision is scarce or unavailable, DRM is the most robust for ground states, while WAN requires careful discriminator-generator balancing and may underperform if energy is learned as a free parameter. In all cases, FBC should be preferred whenever feasible to avoid brittle boundary-loss tuning; OG should be added judiciously and evaluated for degeneracy-induced mixing. These choices can be summarized succinctly by pairing each task with the smallest configuration that achieves the desired \(L_2\) tolerance within the time budget, using the results reported in Tables~\ref{Results for One Dimensional Infinite Potential Well}–\ref{Results for Two Dimensional Quantum Harmonic Oscillator} as reference points.

\section{Threats to Validity and Limitations}

The conclusions are conditioned on the chosen architectures, optimizers, and sampling strategies. Although sensitivity analyses in this chapter indicate that width contributes more than depth for single-network solvers and that most gains appear within the first \num{10000} epochs, other hyperparameter combinations may alter the speed-accuracy frontier. FN requires prior knowledge of nodal locations and therefore lacks generality for arbitrary potentials. OG can entangle degenerate manifolds in higher dimensions unless guided by additional symmetry-aware constraints or targeted initialization. Finally, wall-clock comparisons depend on hardware and implementation details and, while the relative rankings have been consistent across repetitions, absolute times should be interpreted as indicative rather than definitive.

\section{Future Directions}

Several focused extensions can broaden the scope and strengthen the robustness of the present study. On the physics side, a natural progression is from low-dimensional benchmarks to three-dimensional, single-particle Schrödinger problems where analytic or high-precision numerical references enable tight validation of accuracy-cost scaling. Interacting systems are a principled step beyond: two particles in three dimensions yield a six-dimensional configuration space that becomes tractable by separating centre-of-mass and relative coordinates, exploiting symmetries (spin, parity, angular momentum), and embedding those invariances in the architecture or loss. Ultimately, many-body settings remain challenging for classical discretizations; here, neural solvers could act as fast surrogates for reduced models or pair with low-rank and tensor-network parameterizations to control complexity while retaining physical fidelity.

Methodologically, forced boundary conditions (FBC) should remain the default boundary-enforcement mechanism, as they remove brittle loss balancing and concentrate optimization on interior residuals. Time dependence, handled natively by PINNs, can be extended to DRM and WAN. For DRM, the stationary energy-minimization principle generalizes to the time-dependent Schrödinger equation by minimizing an action functional over spacetime (e.g., a least-action or McLachlan variational principle) with temporal boundary or initial conditions imposed by construction; this yields a spacetime deep Ritz formulation that preserves variational structure. For WAN, the weak form can be lifted to spacetime by allowing the discriminator (or test-function network) to integrate over both spatial and temporal domains; periodic or Floquet conditions can be enforced in the KH frame by design. In both cases, careful scaling of temporal and spatial residuals is essential to avoid stiffness, and shared spacetime representations can reduce the need for explicit time marching while remaining mesh-free.

Adaptive, error-driven sampling is a complementary lever. Residual-based refinement, multi-resolution schedules, and simple curricula (e.g., homotopy in potential strength or in the KH parameter~$\alpha$) can concentrate collocation points where they matter most: near sharp potential features, along nodal sets of excited states, and in regions with slow PINN/DRM convergence. Parallel multi-state training with a shared trunk and state-specific heads, coupled to explicit block-orthogonality constraints or a Gram-Schmidt layer, amortizes features across eigenstates and reduces Runtime compared to sequential deflation, while mitigating error propagation from low to high quantum numbers.

The role of fixed/forced nodes (FN) merits deeper study. FN is effective when nodal locations are known and regular (as in the infinite well), but it is brittle for irregular spectra (e.g., the harmonic oscillator) and requires prior solution knowledge. A more general alternative is neural domain decomposition: multiple subnetworks solve on overlapping subdomains with continuity and flux-matching penalties at interfaces (a neural analogue of Schwarz or mortar methods). This preserves the benefits of FBC locally, adapts naturally to complex geometries or unknown nodal patterns, and allows residual-driven refinement of subdomain boundaries. Where node information is partially available, one can jointly learn approximate nodal sets with a small auxiliary predictor and refine them a posteriori using residual indicators, rather than hard-coding fixed nodes.

Two additional strands can enhance practicality. First, operator learning provides amortized solvers over families of potentials, delivering rapid predictions for parameter sweeps and high-quality warm starts for PINN/DRM/WAN fine-tuning on specific instances. Second, stability, efficiency, and credibility benefit from staged optimizers (Adam followed by L–BFGS), discriminator regularization and two-time-scale updates for WAN, symmetry-aware parameterizations (e.g., rotational or gauge equivariance in KH), mixed precision, multi-GPU data parallelism, and systematic uncertainty quantification via ensembles or Bayesian variants alongside residual-based error indicators. Together, these directions aim for robust, scalable neural PDE solvers applicable to higher-dimensional, interacting, and time-dependent quantum systems where classical techniques become costly.

\chapter{Conclusion}
This dissertation investigated three neural PDE solvers, physics-informed neural networks (PINNs), the deep Ritz method (DRM), and weak adversarial networks (WANs), across a broad suite of benchmarks: solving multi-dimensional Poisson problems and multi-state one- and two-dimensional Schr\"odinger equations for the infinite potential well and the quantum harmonic oscillator. Under a unified evaluation protocol, all three approaches achieved low solution errors, typically in the range \(10^{-6}\) to \(10^{-9}\), when complemented by principled design choices such as forced boundary conditions (FBC), fixed or forced nodes (FN) where nodal sets are known, and orthogonality regularization (OG) for reliable recovery of excited states. Building on these foundations, we formulated and solved the laser-driven Schr\"odinger equation in the Kramers-Henneberger (KH) frame, demonstrating that the same methodological toolkit extends to a demanding, time-independent setting and enabling an end-to-end Schr\"odinger solver for this class of problems.

The comparative study yields several clear takeaways. PINN emerges as a dependable, general-purpose choice: with FBC and modest architectural tuning, it consistently delivers strong accuracy across Poisson and Schr\"odinger tasks and scales smoothly to higher excited states. DRM is particularly attractive for stationary problems, combining fast convergence with competitive accuracy; where time dependence is absent, or can be handled through a variational extension, DRM often attains the best accuracy–runtime trade-off. WAN, while more sensitive to capacity balance and training dynamics due to its two-network structure, performs well in regimes where weak-form constraints and FN or OG are leveraged effectively; with careful regularization and scheduling, it achieves accurate solutions on several benchmarks.

Beyond numerical performance, the thesis clarifies why these methods succeed under specific configurations. FBC removes brittle boundary-loss tuning and shifts optimization effort to the interior residual. OG prevents mode collapse in spectral problems by separating eigenstates during training, provided degeneracies are handled with care. FN is powerful when nodal information is reliable and regularly spaced, as in the infinite well, but benefits from more general strategies, such as neural domain decomposition, when nodes are unknown or irregular, as in the harmonic oscillator. Sensitivity analyses and ablations show that most gains accrue within a moderate training budget, that width is usually more impactful than depth for single-network solvers, and that uniform or residual-informed sampling stabilizes training.

There are, of course, limitations. Performance depends on reasonable hyperparameter choices and architectures; degeneracies in higher dimensions can stress OG without symmetry-aware guidance; WAN requires disciplined discriminator regularization. Nonetheless, the evidence across Chapters~4-6 supports a pragmatic recipe: prefer FBC whenever feasible; use PINN as the default for accuracy and for excited spectra; choose DRM where rapid stationary solves are paramount; and deploy WAN when weak-form advantages or FN and OG structure can be exploited. The extension to the KH frame indicates that these principles transfer beyond canonical benchmarks and into driven quantum dynamics.

Looking ahead, several directions appear especially promising: variational and weak-form extensions that bring fully time-dependent DRM and WAN to parity with PINNs, adaptive and residual-driven sampling to focus computation where it matters most, parallel multi-state training with explicit orthogonality constraints to reduce wall-clock runtime, and neural domain decomposition to replace hand-crafted FN in problems with unknown nodal patterns. Operator-learning surrogates, symmetry-aware parameterizations, and systematic uncertainty quantification can further improve scalability, stability, and credibility. Taken together, the results suggest that neural PDE solvers, when anchored by sound physical priors and disciplined training practice, offer a viable and increasingly mature alternative to classical discretizations for complex, high-dimensional scientific problems.

\section*{Acknowledgements}

I would like to express my sincere gratitude to my supervisor, Prof Simon, for his guidance and invaluable advice throughout my research. I am also thankful to Prof Carla who gives the physics guidance of this project.

A special thanks to my parents for their unwavering support and encouragement during my studies.



\bibliographystyle{plain}
\bibliography{references}

\end{document}